%% file: main_iclr.tex
\documentclass{article} 
\usepackage{iclr2023_conference,times}

\input{math_commands.tex}

\usepackage{multirow}
\usepackage{hyperref}
\hypersetup{
    colorlinks=true,
    linkcolor=red,
    filecolor=magenta,      
    urlcolor=purple,
    citecolor=blue,
    pdftitle={Overleaf Example},
    pdfpagemode=FullScreen,
    }
\usepackage{url}
\usepackage{enumitem}
\setlist{nolistsep}
\newcommand{\review}[1]{{\color{black} #1}} 
\newcommand{\rnd}[1]{{\color{black} #1}} 

\usepackage{graphicx}
\usepackage{booktabs}
\usepackage{multirow}
\usepackage{comment}

\title{A Reproducible and Realistic Evaluation of Partial Domain Adaptation Methods}

\author{Tiago Salvador$^{1,2,\dagger}$, Kilian Fatras$^{1,2,\dagger}$, Ioannis Mitliagkas$^{1,3,4}$ \& Adam Oberman$^{1,2,4}$ \\
Mila - Quebec AI Institute$^{1}$, McGill University$^{2}$, Universit\'e de Montr\'eal$^{3}$, Canada CIFAR AI Chair$^{4}$\\
Montr\'eal, Qu\'ebec, Canada \\
}

\iclrfinalcopy 

\begin{document}

\maketitle

\begin{abstract}
Unsupervised Domain Adaptation (UDA) aims at classifying unlabeled target images leveraging source labeled ones. In this work, we consider the Partial Domain Adaptation (PDA) variant, where we have extra source classes not present in the target domain. Most successful algorithms use model selection strategies that rely on target labels to find the best hyper-parameters and/or models along training. However, these strategies violate the main assumption in PDA: only unlabeled target domain samples are available. Moreover, there are also inconsistencies in the experimental settings - architecture, hyper-parameter tuning, number of runs - yielding unfair comparisons. The main goal of this work is to provide a realistic evaluation of PDA methods with the different model selection strategies under a consistent evaluation protocol. We evaluate 7 representative PDA algorithms on 2 different real-world datasets using 7 different model selection strategies. Our two main findings are: \emph{(i)} without target labels for model selection, the accuracy of the methods decreases up to 30 percentage points; \emph{(ii)} only one method and model selection pair performs well on both datasets. Experiments were performed with our PyTorch framework, BenchmarkPDA, which we open source.
\end{abstract}

\section{Introduction}

\let\thefootnote\relax\footnotetext{$^{\dagger}$ denotes equal contributions. Corresponding authors: \{tiago.salvador, kilian.fatras\}@mila.quebec.} 
\let\thefootnote\relax\footnotetext{Preprint.}
{\bfseries Domain adaptation.} Deep neural networks are highly successful in image recognition for in-distribution samples \citep{He16Resnet} with this success being intrinsically tied to the large number of labeled training data. However, they tend to not generalize as well on images with different background or colors not seen during training. Such shift in the samples is referred to as domain shift in the literature. Unfortunately, enriching the training set with new samples from different domains is challenging as labeling data is both an expensive and time-consuming task. {Thus, researchers have focused on unsupervised domain adaptation (UDA) where we have access to unlabelled samples from a different domain, known as the target domain. The purpose of UDA is to classify these unlabeled samples by leveraging the knowledge given by the labeled samples from the source domain \citep{Pan2010, Patel2015}}. In the standard UDA problem, the source and target domains are assumed to share the same classes. In this paper, we consider a more challenging variant of the problem called \emph{partial domain adaptation} (PDA): the classes in the target domain $\mathcal{Y}_t$ form a subset of the classes in the source domain $\mathcal{Y}_s$ \citep{cao2018pada}, \emph{i.e.,} $\mathcal{Y}_t \subset \mathcal{Y}_s$. The number of target classes is unknown as we do not have access to the labels. The extra source classes, not present in the target domain, make the PDA problem more difficult: simply aligning the source and target domains forces a negative transfer where target samples are matched to outlier source-only labels.

{\bfseries Realistic evaluations.} Most recent PDA methods report an increase of the target accuracy up to 15 percentage points on average when compared to the baseline approach that uses only source domain samples. While these successes constitute important breakthroughs in the DA research literature, target labels are used for model selection, violating the main UDA assumption. In their absence, the effectiveness of PDA methods remains unclear and model selection constitutes a yet to be solved problem as we show in this work. Moreover, the hyper-parameter tuning is either unknown or lacks details and sometimes requires labeled target data, which makes it challenging to apply PDA methods to new datasets (see Table \ref{table:sota_summary_exp} for a full summary). Recent work has highlighted the importance of model selection in the presence of domain shift. \citet{gulrajani2021in} showed that when evaluating domain generalization (DG) algorithms, whose goal is to generalize to a completely unseen domain, in a consistent and realistic setting no method outperforms the baseline ERM method by more than 1 percentage point. They argue that DG methods without a model selection strategy remain incomplete and should therefore be specified as part of the method. \citet{saito2021tune} make the same recommendation in the context of UDA and PDA. 

Model selection strategies have been designed in a research environment but they have not been tested extensively with a realistic and fair experimental protocol. In this work, we study their applicability to real-world settings and evaluate 7 different PDA methods with 7 different model selection strategies on 2 different datasets (see Table \ref{table:our_summary_exp} for a summary). We reimplemented all our considered PDA methods with the same architecture, optimizer and learning rate schedule for comparison purposes. We list below our major findings:
\begin{itemize}[leftmargin=5.5mm]
    \setlength\itemsep{0.5em}
    \item The accuracy attained by models selected without target labels can decrease up to 30 percentage points compared to the one reported using target labels (See Table \ref{table:small_overall_results} for a summary of the results). 
    \item Out of the 49 model selection strategies and PDA methods pairs considered, only one gave consistent results over all tasks on both datasets.
    \item Random seed plays an important role in the selection of hyper-parameters. Selected parameters are not stable across different seeds and the standard deviation between accuracies on the same task can be up to $8.4\%$ even when relying on target labels for model selection.
    \item Under a more realistic scenario where some target labels are available, \rnd{100} random samples is enough to see only a drop of 1 percentage point in accuracy (when compared to using all target samples). However, the extreme case of using only one labeled target sample per class leads to significant drop in performance.
\end{itemize}

{\bfseries Outline.} 
In Section \ref{sec:model_selection}, we provide an overview of the different model selection strategies considered in this work. Then in Section \ref{sec:methods}, we discuss the PDA methods that we consider. In Section \ref{sec:exp} we describe the training procedures, hyper-parameter tuning and evaluation protocols used to evaluate all methods fairly. In Section \ref{sec:results}, we discuss the results of the different benchmarked methods and the performance of the different model selection strategies. Finally in Section \ref{sec:conclusion}, we give some recommendations for future work in partial domain adaptation.

\input{tables/table_small_overall_results.tex}

\section{Model Selection Strategies: An Overview}
\label{sec:model_selection}

Model selection (choosing hyper-parameters, training checkpoints, neural network architectures) is a crucial part of training neural networks. In the supervised learning setting, a validation set is used to estimate the model's accuracy. However, in UDA such approach is not possible as we have unlabeled target samples. Several strategies have been designed to address this issue. Below, we discuss the ones used in this work.

{\bfseries Source Accuracy (\textsc{s-acc}).}
\citet{ganin15} used the accuracy estimated on a small validation set from the source domain to perform the model selection. 
While the source and target accuracies are related, there are no theoretical guarantees. \citet{you2019DEV} showed that when the domain gap is large this approach fails to select competitive models.

{\bfseries Deep Embedded Validation (\textsc{dev}).}
\citet{sugiyama2007IWCV} and \citet{LongCADA2018} perform model selection through Importance-Weighted Cross-Validation (\textsc{IWCV}). Under the assumption that the source and target domain follow a covariate shift, the target risk can be estimated from the source risk through importance weights that give increased importance to source samples that are closer to target samples. These importance weights correspond to the ratio of the target and source densities and are estimated using Gaussian kernels. Recently, \citet{you2019DEV} proposed an improved variant, Deep Embedded Validation (\textsc{dev}), that controls the variance of the estimator and estimates the importance weights with a discriminative model that distinguish source samples from target samples leading to a more stable and effective method.

{\bfseries Entropy (\textsc{ent}).} While minimizing the entropy of the target samples has been used in domain adaptation to improve accuracy by promoting tighter clusters, \citet{morerio2018minimalentropy} showed that it can also be used for model selection. The intuition is that a lower entropy model corresponds to a high confident model with discriminative target features and therefore reliable predictions.

{\bfseries Soft Neighborhood Density (\textsc{snd}).}
\citet{saito2021tune} argue that a good UDA model will have a neighborhood structure where nearby target samples are in the same class. They point out that entropy is not able to capture this property and propose the Soft Neighborhood Density (\textsc{snd}) score to address it.

{\bfseries Target Accuracy (\textsc{oracle}).}
We consider as well the target accuracy on all target samples. While we emphasize once again its use is not realistic in unsupervised domain adaptation (hence why we will refer to it as \textsc{oracle}), it has nonetheless been used to report the best accuracy achieved by the model along training in several previous works \citep{cao2018pada, Xu_2019_ICCV, liang2020baus, gu2021adversarial, nguyen2021improving}. Here, we use it as an upper bound for all the other model selection strategies and to check the reproducibility of previous works.

{\bfseries Small Labeled Target Set (\textsc{1-shot} and \textsc{100-rnd}).}
For real-world applications in an industry setting, it is unlikely that a model will be deployed without the very least of an estimate of its performance for which target labels are required. Therefore, one can imagine a situation where a PDA method is used and a small set of target samples is available. Thus, we will compute the target accuracy with 1 labeled sample per class (\textsc{1-shot}) and 100 random labeled target samples (\textsc{100-rnd}) as model selection strategies. One could argue that the 100 random samples could have been used in the training with semi-supervised domain adaptation methods. However, note that we do not know how many classes we have on the target domain so it is hard to form a split when we have uncertainty of classes. For instance, 100 random represents possibly less than 2 samples per class for one of our real-world dataset, as we do not know the number of classes, making a potential split between a train and validation target sets not possible.

\begin{table}[t!]
\centering
\begin{tabular}{l|c|c|c|c}
\multirow{2}{*}{Method} & Architecture & Runs & \multicolumn{2}{c}{Model Selection}\\
 & (bottleneck) & per task & Hyper-Parameters & Along Training\\
\midrule
\textsc{pada}   & Linear     & 1 & \textsc{IWCV} (lacks details) & \textsc{oracle} \\
\textsc{safn}   & Non-Linear & 3 & Unknown & \textsc{oracle}\\
\textsc{ba3us}  & Linear     & 3 & Unknown & \textsc{oracle}\\
\textsc{ar}     & Non-Linear & 1 & \textsc{IWCV} (lacks details) & \textsc{oracle}\\
\textsc{jumbot} & Linear     & 1 & \textsc{oracle} & \textsc{final}\\
\textsc{mpot}   & Linear     & 3 & Unknown   & \textsc{oracle}\\
\end{tabular}
\caption{Summary of the experimental protocol used for SOTA partial domain adaptation methods. We refer to Appendix \ref{appx:benchmarkpda} for additional details.}
\label{table:sota_summary_exp}
\end{table}
\raggedbottom
\section{Partial Domain Adaptation Methods} \label{sec:methods}
In this section, we give a brief description of the PDA methods considered in our study. They can be grouped into two families: adversarial training and divergence minimization.

{\bfseries Adversarial training.}
To solve the UDA problem, \cite{ganin2016domain} aligned the source and target domains with the help of a domain discriminator trained adversarially to be able to distinguish the samples from the two domains. However, when applied to the PDA problem this strategy leads to negative transfer and the model performs worse than a model trained only on source data. \citet{cao2018pada} proposed \textsc{pada} that introduces a PDA specific solution to adversarial domain adaptation: the contribution of the source-only class samples to the training of both the source classifier and the domain adversarial network is decreased. This is achieved through  class weights that are calculated by simply averaging the classifier prediction on all target samples. As the source-only classes should not be predicted in the target domain, they should have lower weights. More recently, \citet{liang2020baus} proposed \textsc{ba3us} which augments the target mini-batch with source samples to transform the PDA problem in a vanilla DA problem. In addition, an adaptive weighted complement entropy objective is used to encourage incorrect classes to have uniform and low prediction scores.

{\bfseries Divergence minimization.} Another standard direction to align the source and target distributions in the feature space of a neural network is to minimize a given divergence between distributions of domains.
\citet{Xu_2019_ICCV} empirically found than target samples have low feature norm compared to source samples. Based on this insight, they proposed \textsc{safn} which progressively adapts the feature norms of the two domains by minimizing the Maximum Mean Feature Norm Discrepancy.
Other approaches are based on optimal transport (OT) \citep{Damodaran_2018_ECCV}. For the PDA problem in specific, \citep{fatras2021minibatch} developed \textsc{jumbot}, a mini-batch unbalanced optimal transport that learns a joint distribution of the embedded samples and labels. The use of unbalanced OT is critical for the PDA problem as it allows to transport only a portion of the mass limiting the negative transfer between distributions. Based on this work, \citep{nguyen2021improving} investigated the partial OT variant \citep{chapel2020partial}, a particular case of unbalanced OT, proposing \textsc{m-pot}. 
Finally, another line of work is to use the Kantorovich-Rubenstein duality of optimal transport to perform the alignment similarly to WGAN\citep{arjovsky17a}. This is precisely the work of \cite{gu2021adversarial} that proposed, \textsc{ar}. In addition,  source samples are reweighted in order to reduce the negative transfer from the source-only class samples. The Kantorovich-Rubenstein duality relies on a one Lipschitz function which is approximated using adversarial training like the PDA methods described above.

\begin{table}[t!]
    \centering
    \begin{tabular}{@{\hskip2pt}c@{\hskip2pt}|c@{\hskip2pt}}
        PDA Methods  & \textsc{pada}, \textsc{safn}, \textsc{ba3us} \textsc{ar}, \textsc{jumbot}, \textsc{mpot}\\
        \midrule
        Model Selection Strategies & \textsc{s-acc}, \textsc{ent}, \textsc{dev}, \textsc{snd}, \textsc{1-shot}, \textsc{100-rnd}, \textsc{oracle} \\
        \midrule
        Architecture & ResNet50 backbone $\oplus$ linear bottleneck $\oplus$ linear classification head\\
        \midrule
        Experimental protocol & 3 seeds on the 12 tasks of \textsc{office-home} and 2 tasks of \textsc{VisDA}\\
    \end{tabular}
    \caption{Summary of the methods, model selection strategies, architecture and datasets considered in this work.}
    \label{table:our_summary_exp}
\end{table}
\raggedbottom
\section{Experimental Protocol}\label{sec:exp}

In this section, we discuss our choices regarding the training details, datasets and  neural network architecture. We then discuss the hyper-parameter tuning used in this work. We summarize the PDA methods, model selection strategies and experimental protocol used in this work in Table \ref{table:our_summary_exp}. The main differences in the experimental protocol of the different published state-of-the-art (SOTA) methods is summarized in Table \ref{table:sota_summary_exp}. To perform our experiments we developed a PyTorch \citep{NEURIPS2019_PyTorch} framework: BenchmarkPDA. We make it available for other researchers to use and contribute with new algorithms and model selection strategies:
\begin{center}
    \url{https://github.com/oberman-lab/BenchmarkPDA}
\end{center}
It is the standard in the literature when proposing a new method to report directly the results of its competitors from the original papers \citep{cao2018pada, Xu_2019_ICCV, liang2020baus, gu2021adversarial, nguyen2021improving}. 
As a result some methods differ for instance in the neural network architecture implementation (\textsc{ar} \citep{gu2021adversarial}, \textsc{safn} \citep{Xu_2019_ICCV}) or evaluation protocol \textsc{jumbot} \citep{fatras2021minibatch} with other methods. These changes often contribute to an increased performance of the newly proposed method leaving previous methods at a disadvantage. Therefore we chose to implement all methods with the same commonly used neural network architecture, optimizer, learning rate schedule and evaluation protocol. We discuss the details below.

\input{tables/table_model_selection_heuristic.tex}

\subsection{Methods, Datasets, Training and Evaluation Details}
{\bfseries Methods.}
We implemented 7 PDA methods by adapting the code from the Official GitHub repositories of each method: Source Only, \textsc{pada} \citep{cao2018pada}, \textsc{safn} \citep{Xu_2019_ICCV}, \textsc{ba3us} \citep{liang2020baus}, \textsc{ar} \citep{gu2021adversarial}, \textsc{jumbot} \citep{fatras2021minibatch}, \textsc{mpot} \citep{nguyen2021improving}. We provide the links to the different official repositories in Appendix \ref{appx:benchmarkpda}. 

{\bfseries Datasets.} We consider two standard real-world datasets used in DA. Our first dataset is \textsc{office-home} \citep{office_home}. It is a difficult dataset for unsupervised domain adaptation (UDA), it has 15,500 images from four different domains: {Art (A), Clipart (C), Product (P) and Real-World (R)}. For each domain, the dataset contains images of 65 object categories that are common in office and home scenarios. For the partial \textsc{office-home} setting, we follow \cite{cao2018pada} and select the first 25 categories (in alphabetic order) in each domain as a partial target domain. We evaluate all methods in all 12 adaptation scenarios.
\textsc{visda} \citep{visda} is a large-scale dataset for UDA. It has 152,397 synthetic images as source domain and 55,388 real-world images as target domain, where 12 object categories are shared by these two domains. For the partial \textsc{VisDA} setting, we follow \cite{cao2018pada} and select the first 6 categories, taken in alphabetic order, in each domain as a partial target domain. We evaluate the models in the two possible scenarios. We highlight that we are the first to investigate the performance of \textsc{jumbot} and \textsc{mpot} on partial \textsc{visda}.

{\bfseries Model Selection Strategies}
We consider the 7 different strategies for model selection described in Section \ref{sec:model_selection}: \textsc{s-acc}, \textsc{dev}, \textsc{ent}, \textsc{snd}, \textsc{oracle}, \textsc{1-shot}, \textsc{100-rnd}. We use them both for hyper-parameter tuning as well selecting the best model along training. Since \textsc{s-acc}, \textsc{dev} and \textsc{snd} require a source validation set, we divide the source samples into a training subset (80\%) and validation subset (20\%). Regardless of the model selection strategy used, all methods are trained using the source training subset. This is in contrast with previous work that uses all source samples, but necessary to ensure a fair comparison of the model selection strategies. We refer to Appendix \ref{appx:model_selection} for additional details.

{\bfseries Architecture.} Our network is composed of a feature extractor with a linear classification layer on top of it. {The feature extractor is a ResNet50 \citep{He16Resnet}, pre-trained on ImageNet \citep{deng2009imagenet}, with its last linear layer removed and replaced by a linear bottleneck layer of dimension 256.}

{\bfseries Optimizer.}
We use the SGD \citep{Robbins1951} algorithm with momentum of 0.9, a weight decay of $5e^{-4}$ and Nesterov acceleration. As the bottleneck and classifer layers are randomly initialized, we set their learning rates to be 10 times that of the pre-trained ResNet50 backbone. We schedule the learning rate with a strategy similar to the one in \citep{ganin2016domain}: $\chi_p = \frac{\chi_0}{(1+\mu i)^{-\nu}}$, where $i$ is the current iteration, $\chi_0 = 0.001$, $\gamma = 0.001$, $\nu = 0.75$. While this schedule is slightly different than the one reported in previous work, it is the one implemented in the different official code implementations. We elaborate in the Appendix \ref{appx:optimizer} on the differences and provide additional details. Finally, as for the mini-batch size, \textsc{jumbot} and \textsc{m-pot} were designed with a stratified sampling, \emph{i.e.,} a balanced source mini-batch with the same number of samples per class. This allows to reduce the negative transfer between domains and is crucial to their success.  On the other hand, it was shown that for some methods (e.g. \textsc{BA3US}) using a larger mini-batch, than what was reported, leads to a decreased performance \citep{fatras2021minibatch}. As a result, we used the default mini-batch strategies for each method. \textsc{jumbot} and \textsc{m-pot} use stratefied mini-batches of size 65 for \textsc{office-home} and 36 for \textsc{visda}. All other methods use a standard random uniform sampling strategy with a mini-batch size of 36.

{\bfseries Evaluation Protocol.}
For the hyper-parameters chosen with each model selection strategy, we run the methods for each task 3 times, each with a different seed (2020, 2021, 2022). We tried to control for the randomness across methods by setting the seeds at the beginning of training. Interestingly, as we discuss in more detail in Section \review{\ref{sec:results}}, some methods demonstrated a non-negligible variance across the different seeds showing that some hyper-parameters and methods are not robust to randomness.w

\input{tables/table_office_home_comparison_reported.tex}

\subsection{Hyper-Parameter Tuning}

Previous works \citep{gulrajani2021in, musgrave2021unsupervised, musgrave2022benchmarking} perform random searches with the same number of runs for each method. In contrast, we perform hyper-parameter grid searches for each method.
As a result, the hyper-parameter tuning budgets differs across the methods depending on the number of hyper-parameters and the chosen grid. While one can argue this leads to an unfair comparison of the methods, in practice in most real-world applications one will be interested in using the best method and our approach will capture precisely that.

The hyper-parameter tuning needs to be performed for each task of each dataset, but that would require a significant computational resources without a clear added benefit. Instead for each dataset, we perform the hyper-parameter tuning on a single task: AC for \textsc{office-home} and SR for \textsc{visda}. This same strategy was adopted in \citep{fatras2021minibatch} and the hyper-parameters were found to generalize to the remaining tasks in the dataset. We conjecture that this may be due to the fact that information regarding the number of target only classes is implicitly hidden in the hyper-parameters. See Appendix \ref{appx:hp} for more details regarding the hyper-parameters and the grids chosen for each method.

Several runs in our hyper-parameter search for \textsc{jumbot}, \textsc{m-pot} and \textsc{ba3us} were unsuccessful with the optimization reaching its end without the model being trained at all. This poses a challenge to \textsc{dev}, \textsc{snd} and \textsc{ent} and its one of the failures modes accounted for in \citep{saito2021tune}. Following their recommendations, for \textsc{jumbot}, \textsc{m-pot} and \textsc{ba3us} we discard runs with low source accuracy. Our threshold was 69.01\% and 89.83\% for the AC task on \textsc{office-home} and TV task on \textsc{visda}, respectively. It corresponds to 90\% of the source accuracy attained by the Source-Only model on each task. We consider 90$\%$ because the ablation study of some methods showed that doing the adaptation decreased slightly the performance on the source domain \citep{Damodaran_2018_ECCV}. See Table \ref{table:model_selection_heuristic} that shows that this heuristic leads to improved results.

Lastly, when choosing the hyper-parameters, we only consider the model at the end of training, discarding the intermediate checkpoint models in order to select hyper-parameters which do not lead to overfitting at the end of training and better generalize to the other tasks. Following the above protocol, for each dataset we trained \emph{468} models in total in order to find the best hyper-parameters. Then, to obtain the results with our neural network architecture on all tasks of each dataset, we trained an additional \emph{1224} models for \textsc{office-home} and \emph{156} models for \textsc{visda}. We additionally trained 231 models with the different neural network architectures for \textsc{ar} and \textsc{safn}. In total, \emph{2547} models were trained to make this study and we present the different results in the next section.

\input{tables/table_overall_results.tex}

\section{Partial domain adaptation experiments}\label{sec:results}

We start the results section by discussing the differences between our reproduced results and the published results from the different PDA methods. Then, we compare the performance of the different model selection strategies. Finally, we discuss the sensitivity of methods to the random seed.

\subsection{Reproducibility Of Previous Results}

We start by ensuring that our reimplementation of PDA methods were done correctly by comparing our reproduced results and the reported results in Table \ref{table:office_home_comparison_reported}. 
On \textsc{office-home}, both \textsc{pada} and \textsc{jumbot} achieved higher average task accuracy (1.6 and 1.7 percentage points, respectively) in our reimplementation, while for \textsc{ba3us} and \textsc{mpot} we recover the reported accuracy in their respective papers. However, we saw a decrease in performance for both \textsc{safn} and \textsc{ar} of roughly 8 and 5 percentage points respectively. This is to be expected due to the differences in the neural network architectures. While we use a linear bottleneck layer, \textsc{safn} uses a nonlinear bottleneck layer. As for \textsc{ar}, they make two significant changes: the linear classification head is replaced by a spherical logistic regression (SLR) layer \citep{Gu2020SLR} and the features are normalized (the 2-norm is set to a dataset dependent value, another hyper-parameter that requires tuning) before feeding them to the classification head. While we account for the first change by comparing to AR (w/ linear) results reported in \citep{gu2021adversarial}, in our neural network architecture we do not normalize the features. These changes, nonlinear bottleneck layer for \textsc{safn} and feature normalization for \textsc{ar}, significantly boost the performance of both methods. When now comparing our reimplementation with the same neural network architectures, our SAFN reimplementation achieves a higher average task accuracy by 3 percentage points, while our AR reimplementation is now only 1 percentage points below. The fact that AR reported results are from only one run, while ours are averaged across 3 distinct seeds, justifies the small remaining gap. Moreover, we report higher accuracy or on par on 4 tasks of the 12 tasks. Given all the above and further discussion of the \textsc{visda} dataset results in Appendix \ref{appx:exp}, our reimplementations are trustworthy and give validity to the results we discuss in the next sections.

\input{tables/table_visda_results_tasks_with_stds.tex}

\subsection{Results for Model Selection Strategies}
\paragraph{Model Selection Strategies (w/ vs w/o target labels)}
All average accuracies on the \textsc{office-home} and \textsc{visda} datasets can be found in Table \ref{table:overall_results}. For all methods on \textsc{office-home}, we can see that the results for model selections strategies which do not use target labels are below the results given by \textsc{oracle}. For some pairs, the drop of performance can be significant, leading some methods to perform on par with the \textsc{s. only} method. That is the case on \textsc{office-home} when \textsc{dev} is paired with either \textsc{ba3us}, \textsc{jumbot} and \textsc{mpot}. Even worse is \textsc{mpot} with \textsc{snd} as the average accuracy is more than 10 percentage points below that of \textsc{s. only} with any model selection strategy. Overall on \textsc{office-home}, except for \textsc{mpot}, all methods when paired with either \textsc{ent} or \textsc{snd} give results that are at most 2 percentage points below compared to when paired with \textsc{oracle}.

A similar situation can be seen over the \textsc{visda} dataset where the accuracy without target labels can be down to 25 percentage points. Yet again, some model selection strategies can lead to scores even worse than \textsc{s. only}. That is the case for \textsc{pada}, \textsc{safn} and \textsc{ba3us}. Contrary to \textsc{office-home}, all model selection strategies without target labels lead to at least one method with results on par or worse in comparison to the \textsc{s. only} method. Overall, no model selection strategy without target labels can lead to score on par to the \textsc{oracle} model selection strategy. Finally, \textsc{pada} performs worse than \textsc{s. only} for most model selection strategies, including the ones which use target labels. However, when combined with \textsc{snd} it performs better than with \textsc{oracle} on average, although still within the standard deviation. This is a consequence of the random seed dependence mentioned before on \textsc{visda}: as the hyper-parameters were chosen by performing just one run, we were simply ``unlucky''. In general, all of this confirms the standard assumption in the literature regarding the difficulty of the \textsc{visda} dataset.

\paragraph{Model Selection Strategies (w/ target labels)}
We recall that the \textsc{oracle} model selection strategy uses all the target samples to compute the accuracy while \textsc{1-shot} and \textsc{100-rnd} use only subsets: \textsc{1-shot} has only one sample per class for a total of 25 and 6 on \textsc{office-home} and \textsc{visda}, respetively, while \textsc{100-rnd} has 100 random target samples. Our results show that using only 100 random target labeled samples is enough to reasonably approximate the target accuracy leading to only a small accuracy drop (one percentage point in almost all cases) for both datasets. Not surprisingly, the gap between the \textsc{1-shot} and \textsc{oracle} model selection strategies is even bigger, leading in some instances to worse results than with a model selection strategy that uses no target labels. This poor performance of the \textsc{1-shot} model selection strategy also highlights that semi-supervised domain adaptation (SSDA) methods are not a straightforward alternative to the \textsc{100-rnd} model selection strategy. While one could argue that the target labels could be leveraged during training like in SSDA methods, one still needs labeled target data to perform model selection. However our results suggest that we would need {at least 3 samples per class} for SSDA methods. In addition, knowing that we have a certain number of labeled samples per class provides information regarding which classes are target only, one of the main assumptions in PDA. In that case, PDA methods could be tweaked. This warrants further study that we leave as future work. Finally, we have also investigated a smaller labeled target set of 50 random samples (\textsc{50-rnd}) instead of 100 random samples. The accuracies of methods using \textsc{50-rnd} were not as good as when using \textsc{100-rnd}. All results of pairs of methods and \textsc{50-rnd} can be found in Appendix \ref{appx:exp}. The smaller performance show that the size of the labeled target set is an important element and we suggest to use at least 100 random samples.

\paragraph{Model Selection Strategies (w/o target labels)}

Overall, only the \textsc{jumbot} and \textsc{snd} pair performed reasonably well with respect to the \textsc{jumbot} and \textsc{oracle} pair on both datasets. All other pairs failed in either one of the datasets. Our experiments show that there is no model selection strategy which performs well for all methods. That is why to deploy models in a real-world scenario, we advise to test selected models on a small labeled target set (our \textsc{100-rnd} model selection strategy) to assess the performance of the models as current model selection without target labels can perform very poorly. 

Our conclusion is that the model selection for PDA methods is still an open problem. We conjecture that it is also the case for all domain adaptation as the considered metrics were developed first for this setting.
For future proposed methods, researchers should specify not only which model selection strategy should be used, but also which hyper-parameter search grid should be considered, in order to deploy them in a real-world scenario.

\subsection{Random Seed Dependence}

Ideally, PDA methods should be robust to the choice of random seed. This is of particular importance when performing hyper-parameter tuning since typically only one run per set of hyper-parameters is done (that was the case in our work as well). We investigate this robustness by averaging all the results presented over three different seeds (2020, 2021 and 2022) and reporting the standard deviations. This is in contrast with previous work where only a single run is reported \citep{fatras2021minibatch,gu2021adversarial}. Other works \citep{cao2018pada, Xu_2019_ICCV, liang2020baus} that report standard deviations do not specify if the random seed is different across runs. 
Results for all tasks on \textsc{visda} dataset are in Table \ref{table:visda_results_tasks_with_stds} and on \textsc{office-home} in Appendix \ref{appx:exp} due to space constraints.

Our experiments show that some methods express a non-negligible instabilities over randomness with respect to any model selection methods. This is particularly true for \textsc{ba3us} when paired with \textsc{dev} and \textsc{1-shot} as model selection strategies: there are several tasks where the standard deviation is above 10\%. While in this case this instability may stem from the poor performance of the model selection strategies, it is also visible when \textsc{oracle} is the model selection strategy used. For instance, the \textsc{m-pot} has a standard deviation of 3.3\% on the AP task of \textsc{office-home} which corresponds to a variance of 11\%. On \textsc{visda} this instability and seed dependence is even larger.

\section{Conclusion}\label{sec:conclusion}

In this paper, we investigated how model selection strategies affect the performance of PDA methods. We performed a quantitative study with seven PDA methods and seven model selection strategies on two real-word datasets. Based on our findings, we provide the following recommendations:

{\textit {i) Target label samples should be used to test models before using them in real-world scenario.}} While this breaks the main PDA assumption, it is {impossible to confidently} deploy PDA models selected without the use of target labels. Indeed, model selection strategies without target labels
lead to a significant drop in performance in most cases 
in comparison to using a small validation set.
We argue that the cost of labeling it outweighs the uncertainty in current model selection strategies.

{\textit {ii) The robustness of new PDA method to randomness should be tested over at least three different seeds.}}
We suggest to use the seeds (2020, 2021, 2022) to allow for a fair comparison with our results.

{\textit {iii) An ablation study should be considered when a novel architecture is proposed to quantify the associated increase of performance.}}

As our work focus on a quantitative study of model selection methods and reproducibility of state-of-the-art partial domain adaptation methods, we do not see any potential ethical concern. Future work will investigate new model selection strategies which can achieve similar results as model selection strategies which use label target samples.

\subsubsection*{Acknowledgments}
This work was partially supported by NSERC Discovery grant (RGPIN-2019-06512) and a Samsung grant. Thanks also to CIFAR for their support through the CIFAR AI Chairs program. Authors thank Christos Tsirigotis and Chen Sun for early comments on the manuscript.



\newpage 

\bibliography{iclr2023_conference}
\bibliographystyle{iclr2023_conference}
\newpage 

\appendix

{\centering{\LARGE\bfseries A Reproducible and Realistic Evaluation of Partial Domain Adaptation Methods}

\vspace{1em}
\centering{{\LARGE\bfseries Supplementary material}}

}
\vspace{2em}

{\bfseries Outline.} The supplementary material of this paper is organized as follows:

\begin{itemize}
    \item In Section \ref{appx:details}, we give more details on our experimental protocol.
    \item In Section \ref{appx:exp}, we provide additional results from our experiments.
\end{itemize}

\section{Additional details on Experimental Protocol}\label{appx:details}

\subsection{Implementations in BenchmarkPDA}\label{appx:benchmarkpda}

In order to reimplement the different PDA methods, we adapted the code from the official repository associated with each of the paper. We list them in Table \ref{table:code_repositories}.

\begin{table}[h]
    \centering
    \resizebox{\textwidth}{!}{
    \begin{tabular}{c|l}
    Method & Code Repository\\
    \midrule
    \textsc{pada} & \scriptsize{\url{https://github.com/thuml/PADA/blob/master/pytorch/src/}}\\
    \textsc{safn} & {\scriptsize{\url{https://github.com/jihanyang/AFN/blob/master/partial/OfficeHome/SAFN/code/}}}\\
    \textsc{ba3us} & {\scriptsize{\url{https://github.com/tim-learn/BA3US/blob/master/}}}\\
    \textsc{ar} & {\scriptsize{\url{https://github.com/XJTU-XGU/Adversarial-Reweighting-for-Partial-Domain-Adaptation}}}\\
    \textsc{jumbot} & {\scriptsize{\url{https://github.com/kilianFatras/JUMBOT}}}\\
    \textsc{m-pot} & {\scriptsize{\url{https://github.com/UT-Austin-Data-Science-Group/Mini-batch-OT/tree/master/PartialDA}}}
    \end{tabular}}
    \caption{Office Github code repositories for the PDA methods considered in this work.}
    \label{table:code_repositories}
\end{table}

One of our main claims regarding previous work is the use of target labels to choose the best model along training. This can be easily verified by inspecting the code. For \textsc{pada} it can be seen on line 240 of the script ``train\_pada.py'', for \textsc{ba3us} in line 116 for the script ``run\_partial.py'', for \textsc{m-pot} it can be seen line 164 of the file ``run\_mOT.py'', for \textsc{safn} it can be seen in the ``eval.py'' file and finally for \textsc{ar} in line 149 of the script ``train.py''.

\subsection{Model Selection}
\label{appx:model_selection}

\textsc{dev} requires learning a discriminative model to distinguish source samples from target samples. Its neural network architecture must be specified as well the training details. \citet{you2019DEV} (\textsc{dev})  use a multilayer perceptron, while \citet{saito2021tune} (\textsc{snd}) use a Support Vector Machine in their reimplementation of \textsc{dev}. We empirically observed the latter to yield more stable weights and so that was the one we used. In order to train the SVM discriminator, following \citep{saito2021tune}, we take 3000 feature embeddings from source samples used in training and 3000 random feature embeddings from target samples, both chosen randomly. We do a 80/20 split into training and test data. The SVM is trained with a linear kernel for a maximum of 4000 iterations. Of 5 different SVM models trained with decay values spaced evenly on log space between $10^{-2}$ and $10^4$ the one that leads to the highest accuracy (in distinguishing source from target features) on the test data split is the chosen one.

As for \textsc{snd}, it also requires specifying a temperature for temperature scaling component of the strategy. We used the default value of 0.05 that is suggested in \citep{saito2021tune}.

Finally, we mention that the samples used for \textsc{100-rnd} were randomly selected and their list is made available together with the code. As for the samples used for \textsc{1-shot}, they are the same as the ones used in semi-supervised domain adaptation.

\subsection{Optimizer}\label{appx:optimizer}

In general, all methods claim to adopt Nesterov's acceleration method as the optimization method with a momentum of 0.9 and setting the weight decay set to $5\times 10^{-4}$. The learning rate follows the annealing strategy as in \cite{ganin2016domain}:
\[
\mu_p = \mu_0 (1 + \alpha · p)^{-\beta},
\]
where $p$ is the training progress linearly changing from 0 to 1, $\mu_0 = 0.01$ and $\alpha = 10$ and $\beta = 0.75$.

However, inspecting the Official code repo for each PDA method, the actual learning schedule is given by
\[
\mu_i = \mu_0 (1 + \alpha · i)^{-\beta},
\]
where $i$ is the iteration number in the training procedure, $\mu_0 = 0.01$ and $\alpha = 0.001$ and $\beta = 0.75$. Only when the total number of iterations is 10000 do the learning rate schedules match. In this work, we followed the latter since it is the one indeed used. For \textsc{office-home}, all methods are trained for 5000 iterations, while for \textsc{visda} they are trained for 10000 iterations, with the exception of the \textsc{s. only} which is trained for 1000 iterations on \textsc{office-home} and 5000 iterations on \textsc{visda}.

\subsection{Hyper-Parameters}\label{appx:hp}

In Table \ref{table:hp_grid}, we report the values used for each hyper-parameter in our grid search. We report in Table \ref{table:hp_chosen} the hyper-parameters chosen by each model selection strategy for each method on both datasets. In addition, for the reproducibility of \textsc{ar} with the proposed architecture in \cite{gu2021adversarial},
a feature normalization layer is added in the bottleneck which requires specifying $r$, the value to which the 2-norm is set. This hyper-parameter is therefore included in the hyper-parameter grid search with the possible values of $[5, 10, 20]$ which are the different values used in the experiments in \citep{gu2021adversarial}.

\input{tables/table_hp_grid.tex}

\input{tables/table_hp_chosen.tex}

\input{tables/table_office_home_results_tasks_with_stds.tex}

\section{Additional Discussion of Results}\label{appx:exp}

In this section, we provide additional results that we could not add to the main paper due to the space constraints.

In Table \ref{tab:office_home_results_tasks_with_stds}, we show the accuracy per task on \textsc{office-home} averaged over three different seeds (2020, 2021, 2022) for all pairs of methods and model selection strategies. 

In Table \ref{table:visda_comparison_reported}, we compare previously reported results with ours on \textsc{visda}. While proposed methods reported results on \textsc{office-home}, only \textsc{pada} and \textsc{ar} results are reported in the original papers for \textsc{visda}. \citet{gu2021adversarial} \textsc{ar}) also report results for \textsc{ba3us}. Analysing the results, we see a 9 percentage point decrease in average task accuracy for \textsc{pada}, but our experiments show that there is a significant seed dependence which we discuss in detail below. This is particularly important since \citet{cao2018pada} (\textsc{pada}) report results from a single run. Comparing our best seeds for \textsc{pada} on the SR and RS tasks, we achieve 58.01\% and 67.9\% accuracy versus a reported 53.53\% and 76.5\%. Moreover, we point out that the official code repository for \textsc{pada} does not include the details to reproduce the \textsc{visda} experiments, so it is possible that minor tweaks (e.g learning rate) are necessary. As for \textsc{ba3us}, our results are within the standard deviation being better on the SR task and worse on the RS task. Finally as for \textsc{ar} we see a decrease in performance which, as the results on \textsc{office-home} show, can be explained by the differences in the neural network architecture.

Finally in Table \ref{tab:overall_results_50}, we show all the average task accuracies from all pairs of methods and model selection strategies on the \textsc{office-home} and \textsc{visda} datasets including the \textsc{50-rnd} model selection strategy.

\input{tables/table_visda_comparison_reported.tex}

\input{tables/table_overral_results_with_50rnd.tex}

\end{document}

%% file: math_commands.tex
\usepackage{amsmath,amsfonts,bm}

\def\eqref#1{equation~\ref{#1}}

\def\1{\bm{1}}

\DeclareMathAlphabet{\mathsfit}{\encodingdefault}{\sfdefault}{m}{sl}
\SetMathAlphabet{\mathsfit}{bold}{\encodingdefault}{\sfdefault}{bx}{n}



%% file: tables/table_small_overall_results.tex
\begin{table}[t!]\centering
\resizebox{\textwidth}{!}{
\begin{tabular}{c|@{\hskip 3pt}c|@{\hskip 3pt}|@{\hskip 3pt}c@{\hskip 3pt}|@{\hskip 3pt}c@{\hskip 3pt}|@{\hskip 3pt}c@{\hskip 3pt}|@{\hskip 3pt}c@{\hskip 3pt}|@{\hskip 3pt}c@{\hskip 3pt}|@{\hskip 3pt}c@{\hskip 3pt}|@{\hskip 3pt}c}
\textsc{Dataset} & Model Selection & \textsc{s. only} & \textsc{pada} & \textsc{safn} & \textsc{ba3us} & \textsc{ar} & \textsc{jumbot} & \textsc{mpot}\\
\midrule\midrule
\multirow{2}{*}{\textsc{office-}} & Worst (w/o target labels) & 59.55 (-2.31) & 52.72 (-11.00) & 61.37 (-1.93) & 62.25 (-13.73) & 64.32 (-8.42) & 61.28 (-15.87) & 46.92 (-30.38)\\
\multirow{2}{*}{\textsc{home}} & Best (w/o target labels) & 60.73 (-1.14) & 63.08 (-0.64) & 62.59 (-0.71) & 75.37 (-0.61) & 70.58 (-2.16) & 74.61 (-2.54) & 66.24 (-11.07)\\
 & \textsc{{oracle}} & 61.87 & 63.72 & 63.30 & 75.98 & 72.73 & 77.15 & 77.31\\
\midrule\midrule
\multirow{2}{*}{\textsc{visda}} & Worst (w/o target labels) & 55.02 (-4.46) & 32.32 (-22.26) & 42.83 (-19.81) & 51.07 (-16.60) & 55.69 (-18.15) & 59.86 (-24.15) & 61.62 (-25.33)\\
 & Best (w/o target labels) & 55.24 (-4.24) & 56.83 (2.26) & 58.62 (-4.02) & 65.58 (-2.09) & 67.20 (-6.65) & 77.69 (-6.31) & 78.40 (-8.54)\\
 & \textsc{{oracle}} & 59.48 & 54.57 & 62.64 & 67.67 & 73.85 & 84.01 & 86.95\\
\end{tabular}}
\caption{Task accuracy average computed over three different seeds (2020, 2021, 2022) on Partial \textsc{office-home} and Partial-\textsc{visda}. For each dataset and PDA method, we display the results of the \emph{worst and best performing model selection that do not use target labels} as well as the \textsc{oracle} model selection strategy. All results can be found in Table \ref{table:overall_results}.}
\label{table:small_overall_results}
\end{table}
\raggedbottom

%% file: tables/table_model_selection_heuristic.tex
\begin{table}[t!]\centering
\resizebox{\textwidth}{!}{
\begin{tabular}{c@{\hskip 2pt}|@{\hskip 2pt}c@{\hskip 2pt}||@{\hskip 2pt}c@{\hskip 2pt}|@{\hskip 2pt}c@{\hskip 2pt}|@{\hskip 2pt}c@{\hskip 2pt}||@{\hskip 2pt}c@{\hskip 2pt}|@{\hskip 2pt}c@{\hskip 2pt}|@{\hskip 2pt}c@{\hskip 2pt}||@{\hskip 2pt}c@{\hskip 2pt}|@{\hskip 2pt}c@{\hskip 2pt}|@{\hskip 2pt}c@{\hskip 2pt}||@{\hskip 2pt}c@{\hskip 2pt}|@{\hskip 2pt}c@{\hskip 2pt}|@{\hskip 2pt}c}
\multirow{2}{*}{Dataset} & \multirow{2}{*}{Variant} & \multicolumn{3}{c||@{\hskip 2pt}}{\textsc{ba3us}} & \multicolumn{3}{c||@{\hskip 2pt}}{\textsc{jumbot}} & \multicolumn{3}{c||@{\hskip 2pt}}{\textsc{mpot}} & \multicolumn{3}{c}{\textsc{safn}}\\
 & & \textsc{ent} & \textsc{dev} & \textsc{snd} & \textsc{ent} & \textsc{dev} & \textsc{snd} & \textsc{ent} & \textsc{dev} & \textsc{snd} & \textsc{ent} & \textsc{dev} & \textsc{snd}\\
\midrule
\multirow{2}{*}{\textsc{office-home}} & Naive & 52.60 & \textbf{63.10} & 44.48 & 52.30 & 26.75 & 17.67 & \textbf{49.01} & 16.72 & \textbf{30.63} & 32.12 & \textbf{49.67} & 5.01\\
 & Heuristic & \textbf{58.45} & \textbf{63.10} & \textbf{60.96} & \textbf{56.24} & \textbf{45.79} & \textbf{55.16} & \textbf{49.01} & \textbf{45.61} & \textbf{30.63} & \textbf{46.27} & \textbf{49.67} & \textbf{49.67}\\
\midrule\midrule
\multirow{2}{*}{\textsc{visda}} & Naive & 39.06 & \textbf{36.99} & 1.14 & 35.89 & \textbf{54.53} & 11.99 & \textbf{75.04} & \textbf{55.33} & 36.11 & \textbf{52.82} & \textbf{53.26} & 0.83\\
 & Heuristic & \textbf{67.50} & 34.94 & \textbf{38.76} & \textbf{47.23} & \textbf{54.53} & \textbf{66.42} & \textbf{75.04} & \textbf{55.33} & \textbf{85.36} & \textbf{52.82} & \textbf{53.26} & \textbf{52.82}\\
\end{tabular}}
\caption{Comparison between the naive model selection strategy and our heuristic approach. Accuracy on AC task for \textsc{office-home} and SR task for \textsc{visda}. Best results in \textbf{bold}.}
\label{table:model_selection_heuristic}
\end{table}

%% file: tables/table_office_home_comparison_reported.tex
\begin{table}[t!]\centering
\resizebox{\textwidth}{!}{
\begin{tabular}{c@{\hskip 3pt}|c@{\hskip 3pt}|c@{\hskip 3pt}|c@{\hskip 3pt}|c@{\hskip 3pt}|c@{\hskip 3pt}|c@{\hskip 3pt}|c@{\hskip 3pt}|c@{\hskip 3pt}|c@{\hskip 3pt}|c@{\hskip 3pt}|c@{\hskip 3pt}|c@{\hskip 3pt}|c}
\textsc{Method} & AC & AP & AR & CA & CP & CR & PA & PC & PR & RA & RC & RP & Avg\\
\midrule
\textsc{s. only}$^\dagger$ & 46.33 & 67.51 & 75.87 & 59.14 & 59.94 & 62.73 & 58.22 & 41.79 & 74.88 & 67.40 & 48.18 & 74.17 & 61.35\\
\textsc{s. only} (Ours) & 45.43 & 68.91 & 79.53 & 55.59 & 57.42 & 65.23 & 59.32 & 40.80 & 75.80 & 69.88 & 47.20 & 77.31 & 61.87\\
\midrule\midrule
\textsc{pada}$^\dagger$ & 51.95 & 67.00 & 78.74 & 52.16 & 53.78 & 59.03 & 52.61 & 43.22 & 78.79 & 73.73 & 56.60 & 77.09 & 62.06\\
\textsc{pada} (Ours) & 50.53 & 67.45 & 80.14 & 57.30 & 54.47 & 64.55 & 61.07 & 40.94 & 79.55 & 73.09 & 54.63 & 80.93 & 63.72\\
\midrule\midrule
\textsc{safn}$^{\dagger *}$ & 58.93 & 76.25 & 81.42 & 70.43 & 72.97 & 77.78 & 72.36 & 55.34 & 80.40 & 75.81 & 60.42 & 79.92 & 71.84\\
\textsc{safn}* (Ours) & 59.98 & 79.85 & 85.18 & 72.02 & 73.73 & 78.54 & 76.09 & 59.32 & 83.25 & 80.04 & 64.20 & 84.44 & 74.72\\
\textsc{safn} (Ours) & 49.57 & 68.55 & 78.26 & 57.91 & 59.29 & 66.81 & 59.87 & 45.29 & 75.98 & 69.08 & 51.68 & 77.29 & 63.30\\
\midrule\midrule
\textsc{ba3us}$^\dagger$ & 60.62 & 83.16 & 88.39 & 71.75 & 72.79 & 83.40 & 75.45 & 61.59 & 86.53 & 79.25 & 62.80 & 86.05 & 75.98\\
\textsc{ba3us} (Ours) & 63.26 & 82.75 & 89.16 & 69.91 & 71.93 & 77.58 & 75.73 & 59.94 & 86.89 & 80.93 & 66.77 & 86.93 & 75.98\\
\midrule\midrule
\textsc{ar}$^{\dagger *}$ & 62.13 & 79.22 & 89.12 & 73.92 & 75.57 & 84.37 & 78.42 & 61.91 & 87.85 & 82.19 & 65.37 & 85.27 & 77.11\\
\textsc{ar}* (Ours) & 62.75 & 81.55 & 89.07 & 71.63 & 73.41 & 82.94 & 75.88 & 61.03 & 85.70 & 79.86 & 62.93 & 85.30 & 76.00\\
\textsc{ar} (Ours) & 57.33 & 79.61 & 86.31 & 69.45 & 71.88 & 79.94 & 70.28 & 53.57 & 83.78 & 77.26 & 59.68 & 83.72 & 72.73\\
\midrule\midrule
\textsc{jumbot}$^\dagger$ & 62.70 & 77.50 & 84.40 & 76.00 & 73.30 & 80.50 & 74.70 & 60.80 & 85.10 & 80.20 & 66.50 & 83.90 & 75.47\\
\textsc{jumbot} (Ours) & 61.87 & 78.19 & 88.11 & 77.69 & 76.75 & 84.15 & 76.83 & 63.72 & 84.80 & 81.79 & 64.70 & 87.17 & 77.15\\
\midrule\midrule
\textsc{mpot}$^\dagger$ & 64.60 & 80.62 & 87.17 & 76.43 & 77.61 & 83.58 & 77.07 & 63.74 & 87.63 & 81.42 & 68.50 & 87.38 & 77.98\\
\textsc{mpot} (Ours) & 64.48 & 80.88 & 86.78 & 76.22 & 77.95 & 82.59 & 75.18 & 64.60 & 84.87 & 80.59 & 67.04 & 86.52 & 77.31\\
\end{tabular}}
\caption{Comparison between reported ($\dagger$) accuracies on partial \textsc{office-home} from published methods with our implementation using the \textsc{oracle} model selection strategy. * denotes different bottleneck architectures.}
\label{table:office_home_comparison_reported}
\end{table}

%% file: tables/table_overall_results.tex
\begin{table}[t!]\centering
\resizebox{\textwidth}{!}{
\begin{tabular}{c|@{\hskip 3pt}c|@{\hskip 3pt}|c@{\hskip 3pt}|c@{\hskip 3pt}|c@{\hskip 3pt}|c@{\hskip 3pt}|c@{\hskip 3pt}|c@{\hskip 3pt}|c}
\textsc{Dataset} & \textsc{Method} & \textsc{s-acc} & \textsc{ent} & \textsc{dev} & \textsc{snd} & \textsc{1-shot} & \textsc{100-rnd} & \textsc{oracle}\\
\midrule\midrule
\multirow{7}{*}{\textsc{office-home}} & \textsc{s. only} & 60.38$\pm$0.5& 60.73$\pm$0.2& 60.22$\pm$0.3& 59.55$\pm$0.3& 58.92$\pm$0.4& 60.34$\pm$0.4& 61.87$\pm$0.3\\
 & \textsc{pada} & 63.08$\pm$0.3& 59.74$\pm$0.5& 52.72$\pm$2.8& 62.36$\pm$0.4& 62.00$\pm$0.5& 63.22$\pm$0.1& 63.72$\pm$0.3\\
 & \textsc{safn} & 62.09$\pm$0.2& 61.37$\pm$0.3& 62.03$\pm$0.4& 62.59$\pm$0.1& 49.30$\pm$0.7& 62.36$\pm$0.2& 63.30$\pm$0.2\\
 & \textsc{ba3us} & \textbf{68.32$\pm$1.1}& 73.36$\pm$0.6& 62.25$\pm$7.1& \textbf{75.37$\pm$0.8}& 65.56$\pm$7.6& 75.19$\pm$0.4& 75.98$\pm$0.3\\
 & \textsc{ar} & 65.68$\pm$0.3& 70.58$\pm$0.4& \textbf{64.32$\pm$0.9}& 70.25$\pm$0.2& 70.56$\pm$0.7& 70.34$\pm$0.2& 72.73$\pm$0.3\\
 & \textsc{jumbot} & 62.89$\pm$0.2& \textbf{74.61$\pm$0.8}& 61.28$\pm$0.1& 72.29$\pm$0.2& \textbf{74.95$\pm$0.1}& \textbf{75.74$\pm$0.3}& 77.15$\pm$0.4\\
 & \textsc{mpot} & 66.24$\pm$0.1& 64.46$\pm$0.1& 61.37$\pm$0.2& 46.92$\pm$0.4& 68.28$\pm$0.2& 73.06$\pm$0.3& \textbf{77.31$\pm$0.5}\\
\midrule\midrule
\multirow{7}{*}{\textsc{visda}} & \textsc{s. only} & 55.15$\pm$2.4& 55.24$\pm$3.2& 55.07$\pm$1.2& 55.02$\pm$2.9& 55.72$\pm$2.2& 58.16$\pm$0.6& 59.48$\pm$0.4\\
 & \textsc{pada} & 47.48$\pm$4.8& 32.32$\pm$4.9& 43.43$\pm$5.3& 56.83$\pm$1.0& 53.15$\pm$2.9& 54.38$\pm$2.7& 54.57$\pm$2.6\\
 & \textsc{safn} & 58.20$\pm$1.7& 42.83$\pm$6.3& 58.62$\pm$1.3& 44.82$\pm$8.8& 56.89$\pm$2.1& 59.09$\pm$2.8& 62.64$\pm$1.5\\
 & \textsc{ba3us} & 55.10$\pm$3.7& 65.58$\pm$1.4& 58.40$\pm$1.4& 51.07$\pm$4.3& 64.77$\pm$1.4& 67.44$\pm$1.2& 67.67$\pm$1.3\\
 & \textsc{ar} & 66.68$\pm$1.0& 64.27$\pm$3.6& \textbf{67.20$\pm$1.5}& 55.69$\pm$0.9& 70.29$\pm$1.7& 72.60$\pm$0.8& 73.85$\pm$0.9\\
 & \textsc{jumbot} & 60.63$\pm$0.7& 62.42$\pm$2.4& 59.86$\pm$0.6& 77.69$\pm$4.2& \textbf{78.34$\pm$1.9}& 83.49$\pm$1.9& 84.01$\pm$1.9\\
 & \textsc{mpot} & \textbf{70.02$\pm$2.0}& \textbf{74.64$\pm$4.4}& 61.62$\pm$1.3& \textbf{78.40$\pm$3.9}& 70.96$\pm$3.7& \textbf{86.69$\pm$5.1}& \textbf{86.95$\pm$5.0}\\
\end{tabular}}
\caption{Task accuracy average over seeds 2020, 2021, 2022 on Partial \textsc{office-home} and Partial \textsc{visda} for the different PDA methods and model selection strategy pairs. Best results in \textbf{bold}.}
\label{table:overall_results}
\end{table}

%% file: tables/table_visda_results_tasks_with_stds.tex
\begin{table}[t!]\centering
\resizebox{\textwidth}{!}{
\begin{tabular}{c@{\hskip 3pt}|c@{\hskip 3pt}|c@{\hskip 3pt}|c@{\hskip 3pt}|c@{\hskip 3pt}|c@{\hskip 3pt}|c@{\hskip 3pt}|c@{\hskip 3pt}|c}
\textsc{Task} & \textsc{Method} & \textsc{s-acc} & \textsc{ent} & \textsc{dev} & \textsc{snd} & \textsc{1-shot} & \textsc{100-rnd} & \textsc{oracle}\\
\midrule\midrule
\multirow{7}{*}{SR} & \textsc{s. only} & 46.96$\:\pm\:$1.5 & 48.17$\:\pm\:$3.9 & 49.00$\:\pm\:$0.9 & 48.17$\:\pm\:$3.9 & 49.43$\:\pm\:$0.8 & 50.01$\:\pm\:$1.6 & 51.86$\:\pm\:$1.4\\
 & \textsc{pada} & 44.56$\:\pm\:$5.9 & 40.83$\:\pm\:$11.3 & 41.04$\:\pm\:$4.3 & 56.14$\:\pm\:$9.7 & 52.94$\:\pm\:$4.3 & 49.34$\:\pm\:$8.4 & 49.34$\:\pm\:$8.4\\
 & \textsc{safn} & 52.04$\:\pm\:$3.5 & 29.86$\:\pm\:$16.7 & 52.42$\:\pm\:$2.9 & 28.46$\:\pm\:$16.5 & 49.97$\:\pm\:$3.3 & 47.83$\:\pm\:$0.6 & 56.88$\:\pm\:$2.1\\
 & \textsc{ba3us} & 44.21$\:\pm\:$3.0 & 71.17$\:\pm\:$1.9 & 48.78$\:\pm\:$1.9 & 46.12$\:\pm\:$7.8 & 66.79$\:\pm\:$1.5 & 71.45$\:\pm\:$0.8 & 71.77$\:\pm\:$1.1\\
 & \textsc{ar} & \textbf{68.39$\:\pm\:$1.3} & 75.28$\:\pm\:$2.9 & \textbf{68.54$\:\pm\:$1.3} & 57.61$\:\pm\:$0.4 & 70.11$\:\pm\:$1.4 & 75.09$\:\pm\:$5.2 & 76.33$\:\pm\:$4.5\\
 & \textsc{jumbot} & 55.23$\:\pm\:$2.3 & 56.25$\:\pm\:$2.1 & 54.35$\:\pm\:$2.0 & 75.23$\:\pm\:$8.4 & \textbf{81.27$\:\pm\:$6.9} & \textbf{89.94$\:\pm\:$1.1} & \textbf{90.55$\:\pm\:$0.5}\\
 & \textsc{mpot} & 64.57$\:\pm\:$2.9 & \textbf{82.10$\:\pm\:$2.0} & 57.02$\:\pm\:$1.5 & \textbf{84.45$\:\pm\:$0.4} & 71.33$\:\pm\:$4.4 & 87.20$\:\pm\:$2.3 & 87.23$\:\pm\:$2.3\\
\midrule\midrule
\multirow{7}{*}{RS} & \textsc{s. only} & 63.34$\:\pm\:$3.4 & 62.32$\:\pm\:$2.7 & 61.13$\:\pm\:$3.3 & 61.88$\:\pm\:$2.3 & 62.00$\:\pm\:$3.9 & 66.30$\:\pm\:$2.0 & 67.11$\:\pm\:$2.1\\
 & \textsc{pada} & 50.39$\:\pm\:$3.8 & 23.80$\:\pm\:$1.6 & 45.82$\:\pm\:$9.2 & 57.53$\:\pm\:$10.3 & 53.36$\:\pm\:$1.7 & 59.43$\:\pm\:$5.8 & 59.81$\:\pm\:$6.2\\
 & \textsc{safn} & 64.37$\:\pm\:$0.7 & 55.80$\:\pm\:$5.2 & 64.82$\:\pm\:$0.5 & 61.19$\:\pm\:$3.3 & 63.82$\:\pm\:$1.0 & 70.34$\:\pm\:$5.8 & 68.40$\:\pm\:$1.2\\
 & \textsc{ba3us} & 65.99$\:\pm\:$4.6 & 59.99$\:\pm\:$1.3 & \textbf{68.01$\:\pm\:$1.9} & 56.01$\:\pm\:$2.9 & 62.75$\:\pm\:$2.6 & 63.44$\:\pm\:$1.9 & 63.56$\:\pm\:$1.8\\
 & \textsc{ar} & 64.97$\:\pm\:$0.8 & 53.26$\:\pm\:$9.7 & 65.86$\:\pm\:$3.5 & 53.78$\:\pm\:$2.1 & 70.46$\:\pm\:$4.7 & 70.11$\:\pm\:$5.0 & 71.36$\:\pm\:$5.5\\
 & \textsc{jumbot} & 66.04$\:\pm\:$1.0 & \textbf{68.59$\:\pm\:$4.6} & 65.36$\:\pm\:$0.8 & \textbf{80.16$\:\pm\:$1.1} & \textbf{75.42$\:\pm\:$4.8} & 77.03$\:\pm\:$2.7 & 77.46$\:\pm\:$3.3\\
 & \textsc{mpot} & \textbf{75.47$\:\pm\:$3.8} & 67.18$\:\pm\:$9.1 & 66.21$\:\pm\:$1.2 & 72.36$\:\pm\:$7.4 & 70.58$\:\pm\:$3.1 & \textbf{86.18$\:\pm\:$8.1} & \textbf{86.67$\:\pm\:$7.8}\\
\midrule\midrule
\multirow{7}{*}{Avg} & \textsc{s. only} & 55.15$\:\pm\:$2.4 & 55.24$\:\pm\:$3.2 & 55.07$\:\pm\:$1.2 & 55.02$\:\pm\:$2.9 & 55.72$\:\pm\:$2.2 & 58.16$\:\pm\:$0.6 & 59.48$\:\pm\:$0.4\\
 & \textsc{pada} & 47.48$\:\pm\:$4.8 & 32.32$\:\pm\:$4.9 & 43.43$\:\pm\:$5.3 & 56.83$\:\pm\:$1.0 & 53.15$\:\pm\:$2.9 & 54.38$\:\pm\:$2.7 & 54.57$\:\pm\:$2.6\\
 & \textsc{safn} & 58.20$\:\pm\:$1.7 & 42.83$\:\pm\:$6.3 & 58.62$\:\pm\:$1.3 & 44.82$\:\pm\:$8.8 & 56.89$\:\pm\:$2.1 & 59.09$\:\pm\:$2.8 & 62.64$\:\pm\:$1.5\\
 & \textsc{ba3us} & 55.10$\:\pm\:$3.7 & 65.58$\:\pm\:$1.4 & 58.40$\:\pm\:$1.4 & 51.07$\:\pm\:$4.3 & 64.77$\:\pm\:$1.4 & 67.44$\:\pm\:$1.2 & 67.67$\:\pm\:$1.3\\
 & \textsc{ar} & 66.68$\:\pm\:$1.0 & 64.27$\:\pm\:$3.6 & \textbf{67.20$\:\pm\:$1.5} & 55.69$\:\pm\:$0.9 & 70.29$\:\pm\:$1.7 & 72.60$\:\pm\:$0.8 & 73.85$\:\pm\:$0.9\\
 & \textsc{jumbot} & 60.63$\:\pm\:$0.7 & 62.42$\:\pm\:$2.4 & 59.86$\:\pm\:$0.6 & 77.69$\:\pm\:$4.2 & \textbf{78.34$\:\pm\:$1.9} & 83.49$\:\pm\:$1.9 & 84.01$\:\pm\:$1.9\\
 & \textsc{mpot} & \textbf{70.02$\:\pm\:$2.0} & \textbf{74.64$\:\pm\:$4.4} & 61.62$\:\pm\:$1.3 & \textbf{78.40$\:\pm\:$3.9} & 70.96$\:\pm\:$3.7 & \textbf{86.69$\:\pm\:$5.1} & \textbf{86.95$\:\pm\:$5.0}\\
\end{tabular}}
\caption{Accuracy of different PDA methods based on different model selection strategies on the 2 Partial \textsc{visda} tasks. Average is done over three seeds (2020, 2021, 2022). Best results in \textbf{bold}.}
\label{table:visda_results_tasks_with_stds}
\end{table}

%% file: tables/table_hp_grid.tex
\begin{table}[]
    \centering
    \begin{tabular}{c|c|c}
    Method & HP & Values\\
    \midrule
    \textsc{pada} & $\lambda$ & $[0.1, 0.5, 1.0, 5.0, 10.0]$\\
    \midrule\midrule
    \multirow{2}{*}{\textsc{ba3us}} & $\lambda_{wce}$ & $[0.1, 0.5, 1, 5, 10]$\\
     & $\lambda_{ent}$ & $[0.01, 0.05, 0.1, 0.5, 1]$\\
    \midrule\midrule
    \multirow{2}{*}{\textsc{safn}} & $\lambda$ & $[0.005, 0.01, 0.05, 0.1, 0.5]$\\
     & $\Delta_r$ & $[0.01, 0.1, 1.0]$\\
    \midrule\midrule
    \multirow{4}{*}{\textsc{ar}} & $\rho_0$ & $[2.5, 5.0, 7.5, 10.0]$\\
     & $A_{up}$ & $[5.0, 10.0]$\\
     & $A_{low}$ & $-A_{up}$\\
     & $\lambda_{ent}$ & $[0.01, 0.1, 1.0]$\\
    \midrule\midrule
    \multirow{4}{*}{\textsc{jumbot}} & $\tau$ & $[0.001, 0.01, 0.1]$\\
     & $\eta_1$ & $[0.00001, 0.0001, 0.001, 0.01, 0.1]$\\
     & $\eta_2$ & $[0.1, 0.5, 1.]$\\
     & $\eta_3$ & $[5, 10, 20]$\\
    \midrule\midrule
    \multirow{4}{*}{\textsc{mpot}} & $\epsilon$ & $[0.5, 1.0, 1.5]$\\
     & $\eta_1$ & $[0.0001, 0.001, 0.01, 0.1, 1.0]$\\
     & $\eta_2$ & $[0.1, 1.0, 5.0, 10.0]$\\
     & $m$ & $[0.1, 0.2, 0.3, 0.4]$\\
    \end{tabular}
    \caption{Hyper-Parameter values for each PDA method considered in the grid search.}
    \label{table:hp_grid}
\end{table}

%% file: tables/table_hp_chosen.tex
\begin{table}[h]\centering 
\resizebox{\textwidth}{!}{
\begin{tabular}{c@{\hskip 2pt}|@{\hskip 2pt}c@{\hskip 2pt}|@{\hskip 2pt}c@{\hskip 2pt}|@{\hskip 2pt}c@{\hskip 2pt}|@{\hskip 2pt}c@{\hskip 2pt}|@{\hskip 2pt}c@{\hskip 2pt}|@{\hskip 2pt}c@{\hskip 2pt}|@{\hskip 2pt}c@{\hskip 2pt}|@{\hskip 2pt}c@{\hskip 2pt}|@{\hskip 2pt}c@{\hskip 2pt}|@{\hskip 2pt}c}
Method & Dataset & HP & \textsc{oracle} & \textsc{1-shot} & \textsc{50-rnd} & \textsc{100-rnd} & \textsc{s-acc} & \textsc{ent} & \textsc{dev} & \textsc{snd}\\
\midrule\midrule
\multirow{2}{*}{\textsc{pada}} & \textsc{office-home} & $\lambda$ & 0.5 & 0.1 & 0.1 & 0.5 & 0.1 & 1.0 & 5.0 & 0.5\\
\cmidrule[0.5pt](l{-0.5ex}){2-11}
 & \textsc{visda} & $\lambda$ & 0.5 & 1.0 & 10.0 & 0.5 & 1.0 & 0.5 & 5.0 & 0.1\\
\midrule\midrule
\multirow{4}{*}{\textsc{safn}} & \multirow{2}{*}{\textsc{office-home}} & $\lambda$ & 0.005 & 0.1 & 0.005 & 0.01 & 0.005 & 0.01 & 0.005 & 0.005\\
 & & $\Delta r$ & 0.1 & 0.01 & 0.01 & 0.01 & 0.01 & 0.1 & 0.1 & 0.1\\
\cmidrule[0.5pt](l{-0.5ex}){2-11}
 & \multirow{2}{*}{\textsc{visda}} & $\lambda$ & 0.005 & 0.005 & 0.05 & 0.05 & 0.005 & 0.05 & 0.005 & 0.05\\
 & & $\Delta r$ & 0.1 & 0.01 & 0.01 & 0.01 & 0.01 & 0.01 & 0.01 & 0.01\\
\midrule\midrule
\multirow{4}{*}{\textsc{ba3us}} & \multirow{2}{*}{\textsc{office-home}} & $\lambda_{wce}$ & 5.0 & 10.0 & 5.0 & 5.0 & 5.0 & 0.1 & 10.0 & 1.0\\
 & & $\lambda_{ent}$ & 0.05 & 0.05 & 0.01 & 0.05 & 0.01 & 0.1 & 0.05 & 0.01\\
\cmidrule[0.5pt](l{-0.5ex}){2-11}
 & \multirow{2}{*}{\textsc{visda}} & $\lambda_{wce}$ & 1.0 & 1.0 & 0.1 & 1.0 & 5.0 & 1.0 & 5.0 & 5.0\\
 & & $\lambda_{ent}$ & 0.5 & 0.5 & 0.5 & 0.5 & 0.05 & 0.5 & 0.05 & 1.0\\
\midrule\midrule
\multirow{8}{*}{\textsc{ar}} & \multirow{4}{*}{\textsc{office-home}} & $\rho_0$ & 2.5 & 2.5 & 5.0 & 5.0 & 2.5 & 5.0 & 7.5 & 10.0\\
 & & $A_{up}$ & 5.0 & 5.0 & 10.0 & 5.0 & 5.0 & 10.0 & 10.0 & 10.0\\
 & & $A_{low}$ & -5.0 & -5.0 & -10.0 & -5.0 & -5.0 & -10.0 & -10.0 & -10.0\\
 & & $\lambda_{ent}$ & 0.1 & 0.1 & 1.0 & 1.0 & 0.01 & 1.0 & 0.01 & 1.0\\
\cmidrule[0.5pt](l{-0.5ex}){2-11}
 & \multirow{4}{*}{\textsc{visda}} & $\rho_0$ & 2.5 & 2.5 & 2.5 & 2.5 & 2.5 & 7.5 & 2.5 & 10.0\\
 & & $A_{up}$ & 10.0 & 10.0 & 10.0 & 10.0 & 5.0 & 10.0 & 10.0 & 10.0\\
 & & $A_{low}$ & -10.0 & -10.0 & -10.0 & -10.0 & -5.0 & -10.0 & -10.0 & -10.0\\
 & & $\lambda_{ent}$ & 0.1 & 0.1 & 0.1 & 0.1 & 0.01 & 0.1 & 0.01 & 0.01\\
\midrule\midrule
\multirow{8}{*}{\textsc{jumbot}} & \multirow{4}{*}{\textsc{office-home}} & $\tau$ & 0.01 & 0.01 & 0.01 & 0.001 & 0.1 & 0.01 & 0.01 & 0.001\\
 & & $\eta_1$ & 0.0001 & 0.0001 & 0.001 & 0.0001 & 0.01 & 1e-05 & 0.01 & 1e-05\\
 & & $\eta_2$ & 0.5 & 1.0 & 0.5 & 0.1 & 0.1 & 0.5 & 1.0 & 1.0\\
 & & $\eta_3$ & 10.0 & 5.0 & 5.0 & 5.0 & 5.0 & 20.0 & 10.0 & 5.0\\
\cmidrule[0.5pt](l{-0.5ex}){2-11}
 & \multirow{4}{*}{\textsc{visda}} & $\tau$ & 0.01 & 0.01 & 0.01 & 0.01 & 0.001 & 0.01 & 0.001 & 0.01\\
 & & $\eta_1$ & 0.001 & 0.001 & 0.001 & 0.001 & 0.01 & 1e-05 & 0.01 & 0.0001\\
 & & $\eta_2$ & 1.0 & 1.0 & 0.5 & 1.0 & 0.1 & 0.5 & 1.0 & 1.0\\
 & & $\eta_3$ & 5.0 & 5.0 & 5.0 & 5.0 & 10.0 & 5.0 & 20.0 & 5.0\\
\midrule\midrule
\multirow{8}{*}{\textsc{mpot}} & \multirow{4}{*}{\textsc{office-home}} & $\epsilon$ & 0.5 & 0.5 & 1.0 & 0.5 & 1.0 & 1.5 & 1.0 & 1.5\\
 & & $\eta_1$ & 0.01 & 0.01 & 0.01 & 0.01 & 0.001 & 0.0001 & 1.0 & 0.01\\
 & & $\eta_2$ & 10.0 & 1.0 & 1.0 & 1.0 & 1.0 & 10.0 & 0.1 & 1.0\\
 & & $m$ & 0.3 & 0.1 & 0.1 & 0.2 & 0.3 & 0.4 & 0.2 & 0.4\\
\cmidrule[0.5pt](l{-0.5ex}){2-11}
 & \multirow{4}{*}{\textsc{visda}} & $\epsilon$ & 0.5 & 0.5 & 0.5 & 0.5 & 1.0 & 1.0 & 1.0 & 0.5\\
 & & $\eta_1$ & 0.01 & 0.001 & 0.01 & 0.01 & 0.001 & 0.0001 & 0.0001 & 0.01\\
 & & $\eta_2$ & 1.0 & 1.0 & 1.0 & 1.0 & 1.0 & 10.0 & 1.0 & 10.0\\
 & & $m$ & 0.3 & 0.1 & 0.3 & 0.3 & 0.2 & 0.4 & 0.2 & 0.3\\
\end{tabular}}
\caption{Hyper-parameters selected for the different methods for each model selection strategy on both \textsc{office-home} and \textsc{visda}.}
\label{table:hp_chosen}
\end{table}

%% file: tables/table_office_home_results_tasks_with_stds.tex
\begin{table}[t!]\centering
\resizebox{\textwidth}{!}{
\begin{tabular}{c@{\hskip 3pt}|@{\hskip 3pt}c@{\hskip 3pt}|@{\hskip 3pt}c@{\hskip 3pt}|@{\hskip 3pt}c@{\hskip 3pt}|@{\hskip 3pt}c@{\hskip 3pt}|@{\hskip 3pt}c@{\hskip 3pt}|@{\hskip 3pt}c@{\hskip 3pt}|@{\hskip 3pt}c@{\hskip 3pt}|@{\hskip 3pt}c@{\hskip 3pt}|@{\hskip 3pt}c@{\hskip 3pt}|@{\hskip 3pt}c@{\hskip 3pt}|@{\hskip 3pt}c@{\hskip 3pt}|@{\hskip 3pt}c@{\hskip 3pt}|@{\hskip 3pt}c@{\hskip 3pt}|@{\hskip 3pt}c}
\textsc{Metric} & \textsc{Method} & AC & AP & AR & CA & CP & CR & PA & PC & PR & RA & RC & RP & Avg\\
\midrule\midrule
\multirow{7}{*}{\textsc{s-acc}} & \textsc{s. only} & 44.50$\:\pm\:$1.7 & 67.71$\:\pm\:$2.4 & 78.37$\:\pm\:$0.3 & 52.56$\:\pm\:$0.9 & 54.81$\:\pm\:$0.1 & 62.88$\:\pm\:$0.9 & 58.77$\:\pm\:$0.5 & 39.28$\:\pm\:$0.8 & 75.08$\:\pm\:$0.5 & 68.90$\:\pm\:$0.6 & 45.33$\:\pm\:$1.0 & 76.34$\:\pm\:$0.7 & 60.38$\:\pm\:$0.5\\
 & \textsc{pada} & 50.15$\:\pm\:$2.8 & 66.93$\:\pm\:$1.2 & 76.73$\:\pm\:$1.7 & 58.00$\:\pm\:$1.4 & 56.13$\:\pm\:$1.4 & 66.45$\:\pm\:$0.8 & 60.33$\:\pm\:$2.1 & 43.50$\:\pm\:$1.2 & 76.70$\:\pm\:$0.4 & 69.27$\:\pm\:$3.5 & 53.93$\:\pm\:$1.3 & 78.88$\:\pm\:$0.8 & 63.08$\:\pm\:$0.3\\
 & \textsc{safn} & 47.36$\:\pm\:$0.1 & 66.82$\:\pm\:$1.9 & 77.62$\:\pm\:$0.2 & 57.85$\:\pm\:$0.6 & 57.89$\:\pm\:$0.7 & 66.92$\:\pm\:$0.9 & 58.80$\:\pm\:$0.7 & 42.49$\:\pm\:$0.6 & 75.46$\:\pm\:$0.4 & 67.92$\:\pm\:$0.0 & 49.73$\:\pm\:$0.1 & 76.23$\:\pm\:$0.8 & 62.09$\:\pm\:$0.2\\
 & \textsc{ba3us} & \textbf{54.89$\:\pm\:$4.7} & 71.34$\:\pm\:$0.8 & \textbf{81.91$\:\pm\:$3.9} & 61.68$\:\pm\:$5.2 & \textbf{67.13$\:\pm\:$3.9} & \textbf{72.96$\:\pm\:$1.0} & \textbf{68.90$\:\pm\:$5.0} & \textbf{55.92$\:\pm\:$1.3} & \textbf{79.13$\:\pm\:$4.7} & \textbf{72.27$\:\pm\:$3.5} & 51.84$\:\pm\:$0.5 & \textbf{81.85$\:\pm\:$4.1} & \textbf{68.32$\:\pm\:$1.1}\\
 & \textsc{ar} & 51.12$\:\pm\:$1.2 & \textbf{72.79$\:\pm\:$0.7} & 77.91$\:\pm\:$0.2 & \textbf{63.21$\:\pm\:$1.5} & 60.54$\:\pm\:$4.0 & 72.76$\:\pm\:$0.9 & 63.39$\:\pm\:$3.1 & 48.36$\:\pm\:$1.7 & 78.02$\:\pm\:$1.7 & 70.00$\:\pm\:$1.1 & 52.52$\:\pm\:$1.0 & 77.55$\:\pm\:$2.6 & 65.68$\:\pm\:$0.3\\
 & \textsc{jumbot} & 49.07$\:\pm\:$0.2 & 65.45$\:\pm\:$0.4 & 77.14$\:\pm\:$0.3 & 60.09$\:\pm\:$0.1 & 59.59$\:\pm\:$1.3 & 66.67$\:\pm\:$1.3 & 60.24$\:\pm\:$1.0 & 43.60$\:\pm\:$0.0 & 74.43$\:\pm\:$0.9 & 70.19$\:\pm\:$0.5 & 51.12$\:\pm\:$1.1 & 77.12$\:\pm\:$1.3 & 62.89$\:\pm\:$0.2\\
 & \textsc{mpot} & 53.07$\:\pm\:$0.3 & 72.61$\:\pm\:$1.2 & 78.50$\:\pm\:$0.7 & 61.92$\:\pm\:$0.5 & 64.16$\:\pm\:$1.8 & 70.22$\:\pm\:$0.2 & 64.13$\:\pm\:$0.9 & 50.87$\:\pm\:$1.1 & 77.40$\:\pm\:$0.1 & 70.40$\:\pm\:$0.6 & \textbf{53.99$\:\pm\:$1.5} & 77.61$\:\pm\:$0.3 & 66.24$\:\pm\:$0.1\\
\midrule\midrule
\multirow{7}{*}{\textsc{ent}} & \textsc{s. only} & 45.27$\:\pm\:$1.1 & 68.91$\:\pm\:$1.4 & 79.26$\:\pm\:$0.7 & 54.21$\:\pm\:$2.1 & 55.52$\:\pm\:$0.6 & 63.19$\:\pm\:$0.3 & 56.96$\:\pm\:$1.5 & 38.75$\:\pm\:$0.6 & 75.65$\:\pm\:$1.3 & 69.24$\:\pm\:$1.0 & 45.31$\:\pm\:$1.0 & 76.47$\:\pm\:$0.8 & 60.73$\:\pm\:$0.2\\
 & \textsc{pada} & 46.03$\:\pm\:$2.9 & 62.09$\:\pm\:$2.8 & 76.05$\:\pm\:$1.4 & 55.07$\:\pm\:$2.7 & 47.28$\:\pm\:$0.1 & 60.92$\:\pm\:$2.4 & 56.69$\:\pm\:$2.8 & 38.43$\:\pm\:$3.0 & 77.08$\:\pm\:$0.2 & 69.48$\:\pm\:$1.3 & 49.73$\:\pm\:$3.5 & 78.00$\:\pm\:$1.7 & 59.74$\:\pm\:$0.5\\
 & \textsc{safn} & 47.08$\:\pm\:$2.0 & 66.83$\:\pm\:$0.5 & 77.73$\:\pm\:$0.2 & 56.54$\:\pm\:$2.2 & 59.07$\:\pm\:$0.7 & 66.22$\:\pm\:$0.5 & 56.75$\:\pm\:$2.1 & 39.58$\:\pm\:$2.0 & 73.90$\:\pm\:$0.9 & 67.80$\:\pm\:$0.2 & 48.76$\:\pm\:$0.1 & 76.23$\:\pm\:$0.7 & 61.37$\:\pm\:$0.3\\
 & \textsc{ba3us} & \textbf{59.26$\:\pm\:$0.9} & 76.38$\:\pm\:$1.5 & \textbf{86.03$\:\pm\:$0.6} & 68.96$\:\pm\:$1.8 & 71.07$\:\pm\:$0.8 & 76.22$\:\pm\:$1.2 & \textbf{73.16$\:\pm\:$0.6} & 57.91$\:\pm\:$2.5 & \textbf{85.59$\:\pm\:$1.2} & 78.11$\:\pm\:$1.4 & \textbf{62.85$\:\pm\:$2.7} & 84.84$\:\pm\:$0.6 & 73.36$\:\pm\:$0.6\\
 & \textsc{ar} & 54.91$\:\pm\:$1.8 & \textbf{78.45$\:\pm\:$1.8} & 84.23$\:\pm\:$0.9 & 64.86$\:\pm\:$2.3 & 68.16$\:\pm\:$3.5 & \textbf{80.45$\:\pm\:$0.8} & 67.58$\:\pm\:$0.4 & 52.34$\:\pm\:$1.0 & 82.48$\:\pm\:$1.9 & 74.75$\:\pm\:$2.1 & 55.64$\:\pm\:$1.2 & 83.06$\:\pm\:$1.2 & 70.58$\:\pm\:$0.4\\
 & \textsc{jumbot} & 57.69$\:\pm\:$5.6 & 75.44$\:\pm\:$1.4 & 85.24$\:\pm\:$2.7 & \textbf{75.97$\:\pm\:$1.4} & \textbf{74.85$\:\pm\:$3.3} & 79.75$\:\pm\:$1.2 & 72.85$\:\pm\:$2.4 & \textbf{60.18$\:\pm\:$0.9} & 83.21$\:\pm\:$1.1 & \textbf{81.97$\:\pm\:$1.0} & 61.81$\:\pm\:$4.6 & \textbf{86.33$\:\pm\:$1.6} & \textbf{74.61$\:\pm\:$0.8}\\
 & \textsc{mpot} & 52.94$\:\pm\:$2.0 & 68.94$\:\pm\:$1.2 & 75.98$\:\pm\:$0.6 & 60.58$\:\pm\:$0.8 & 65.99$\:\pm\:$2.2 & 71.51$\:\pm\:$0.8 & 58.28$\:\pm\:$0.9 & 49.87$\:\pm\:$2.6 & 73.77$\:\pm\:$1.3 & 64.98$\:\pm\:$0.4 & 57.53$\:\pm\:$0.6 & 73.17$\:\pm\:$2.7 & 64.46$\:\pm\:$0.1\\
\midrule\midrule
\multirow{7}{*}{\textsc{dev}} & \textsc{s. only} & 43.74$\:\pm\:$1.8 & 67.81$\:\pm\:$1.2 & 78.28$\:\pm\:$0.7 & 51.42$\:\pm\:$2.7 & 54.55$\:\pm\:$1.2 & 63.94$\:\pm\:$1.7 & 57.94$\:\pm\:$0.9 & 39.40$\:\pm\:$0.9 & 74.91$\:\pm\:$0.6 & 69.27$\:\pm\:$1.0 & 45.33$\:\pm\:$1.0 & 75.99$\:\pm\:$1.3 & 60.22$\:\pm\:$0.3\\
 & \textsc{pada} & 44.70$\:\pm\:$1.3 & 61.61$\:\pm\:$5.4 & 68.99$\:\pm\:$11.3 & 35.08$\:\pm\:$13.1 & 24.24$\:\pm\:$20.5 & 61.66$\:\pm\:$2.4 & 57.91$\:\pm\:$1.7 & 38.03$\:\pm\:$0.6 & 73.11$\:\pm\:$3.4 & 66.33$\:\pm\:$0.7 & 29.97$\:\pm\:$21.0 & 71.07$\:\pm\:$11.3 & 52.72$\:\pm\:$2.8\\
 & \textsc{safn} & 48.12$\:\pm\:$0.4 & 67.30$\:\pm\:$0.5 & 77.43$\:\pm\:$0.5 & 56.75$\:\pm\:$0.3 & 58.17$\:\pm\:$1.2 & 65.64$\:\pm\:$1.3 & 59.08$\:\pm\:$0.5 & 43.00$\:\pm\:$1.1 & 74.64$\:\pm\:$0.4 & 68.11$\:\pm\:$0.9 & 50.53$\:\pm\:$0.6 & 75.65$\:\pm\:$0.5 & 62.03$\:\pm\:$0.4\\
 & \textsc{ba3us} & 41.67$\:\pm\:$18.9 & 50.05$\:\pm\:$28.7 & 63.74$\:\pm\:$26.1 & 60.70$\:\pm\:$2.2 & 59.08$\:\pm\:$10.9 & 67.88$\:\pm\:$0.9 & \textbf{64.62$\:\pm\:$1.6} & \textbf{56.74$\:\pm\:$1.3} & 75.21$\:\pm\:$0.6 & \textbf{70.92$\:\pm\:$2.0} & \textbf{58.39$\:\pm\:$2.3} & \textbf{78.06$\:\pm\:$1.3} & 62.25$\:\pm\:$7.1\\
 & \textsc{ar} & \textbf{49.25$\:\pm\:$2.8} & \textbf{70.20$\:\pm\:$1.7} & \textbf{79.73$\:\pm\:$2.5} & \textbf{62.72$\:\pm\:$1.0} & \textbf{61.85$\:\pm\:$4.6} & \textbf{70.86$\:\pm\:$5.6} & 61.65$\:\pm\:$1.0 & 43.72$\:\pm\:$0.7 & \textbf{76.29$\:\pm\:$0.7} & 70.31$\:\pm\:$1.7 & 49.61$\:\pm\:$0.8 & 75.61$\:\pm\:$0.4 & \textbf{64.32$\:\pm\:$0.9}\\
 & \textsc{jumbot} & 46.11$\:\pm\:$0.1 & 66.33$\:\pm\:$0.6 & 76.42$\:\pm\:$0.3 & 56.81$\:\pm\:$0.1 & 56.36$\:\pm\:$0.5 & 66.70$\:\pm\:$0.8 & 58.03$\:\pm\:$1.1 & 41.99$\:\pm\:$0.8 & 74.97$\:\pm\:$0.5 & 67.43$\:\pm\:$0.3 & 48.12$\:\pm\:$0.5 & 76.04$\:\pm\:$0.1 & 61.28$\:\pm\:$0.1\\
 & \textsc{mpot} & 46.07$\:\pm\:$0.7 & 65.43$\:\pm\:$0.8 & 76.46$\:\pm\:$0.4 & 56.44$\:\pm\:$1.0 & 57.95$\:\pm\:$1.0 & 66.35$\:\pm\:$1.0 & 57.64$\:\pm\:$0.8 & 43.60$\:\pm\:$0.6 & 74.86$\:\pm\:$1.3 & 67.68$\:\pm\:$0.5 & 48.12$\:\pm\:$0.8 & 75.89$\:\pm\:$0.4 & 61.37$\:\pm\:$0.2\\
\midrule\midrule
\multirow{7}{*}{\textsc{snd}} & \textsc{s. only} & 42.23$\:\pm\:$1.3 & 68.91$\:\pm\:$1.4 & 79.35$\:\pm\:$0.6 & 51.76$\:\pm\:$3.7 & 53.48$\:\pm\:$2.1 & 63.94$\:\pm\:$1.7 & 55.37$\:\pm\:$0.6 & 37.35$\:\pm\:$1.0 & 74.10$\:\pm\:$2.8 & 68.53$\:\pm\:$1.4 & 43.78$\:\pm\:$0.6 & 75.84$\:\pm\:$1.6 & 59.55$\:\pm\:$0.3\\
 & \textsc{pada} & 50.43$\:\pm\:$0.8 & 66.72$\:\pm\:$1.5 & 79.72$\:\pm\:$1.8 & 57.30$\:\pm\:$1.9 & 52.10$\:\pm\:$1.7 & 63.11$\:\pm\:$1.9 & 60.82$\:\pm\:$3.0 & 39.26$\:\pm\:$2.0 & 79.33$\:\pm\:$1.3 & 73.09$\:\pm\:$1.5 & 45.77$\:\pm\:$1.6 & 80.62$\:\pm\:$0.4 & 62.36$\:\pm\:$0.4\\
 & \textsc{safn} & 49.57$\:\pm\:$0.3 & 68.18$\:\pm\:$1.3 & 77.86$\:\pm\:$0.5 & 57.91$\:\pm\:$0.3 & 58.17$\:\pm\:$1.2 & 66.13$\:\pm\:$1.0 & 59.14$\:\pm\:$0.8 & 43.90$\:\pm\:$0.5 & 75.81$\:\pm\:$0.7 & 68.17$\:\pm\:$1.6 & 49.59$\:\pm\:$1.6 & 76.64$\:\pm\:$0.5 & 62.59$\:\pm\:$0.1\\
 & \textsc{ba3us} & \textbf{62.21$\:\pm\:$0.9} & \textbf{83.29$\:\pm\:$0.4} & \textbf{88.50$\:\pm\:$0.6} & 68.50$\:\pm\:$0.9 & 71.45$\:\pm\:$3.6 & 76.96$\:\pm\:$0.6 & \textbf{76.19$\:\pm\:$1.2} & \textbf{59.94$\:\pm\:$1.7} & \textbf{86.31$\:\pm\:$1.4} & 79.46$\:\pm\:$1.4 & \textbf{65.35$\:\pm\:$1.9} & 86.35$\:\pm\:$0.9 & \textbf{75.37$\:\pm\:$0.8}\\
 & \textsc{ar} & 54.37$\:\pm\:$1.6 & 79.01$\:\pm\:$2.2 & 84.54$\:\pm\:$0.8 & 64.52$\:\pm\:$1.6 & 68.05$\:\pm\:$3.2 & 79.16$\:\pm\:$2.8 & 65.60$\:\pm\:$1.7 & 51.28$\:\pm\:$1.6 & 83.05$\:\pm\:$1.1 & 75.02$\:\pm\:$1.6 & 55.02$\:\pm\:$1.8 & 83.40$\:\pm\:$0.9 & 70.25$\:\pm\:$0.2\\
 & \textsc{jumbot} & 56.60$\:\pm\:$2.8 & 68.48$\:\pm\:$1.5 & 84.70$\:\pm\:$2.1 & \textbf{71.81$\:\pm\:$1.8} & \textbf{71.84$\:\pm\:$1.6} & \textbf{80.91$\:\pm\:$0.9} & 70.28$\:\pm\:$0.8 & 50.69$\:\pm\:$4.9 & 83.89$\:\pm\:$1.5 & \textbf{81.21$\:\pm\:$0.6} & 58.85$\:\pm\:$1.7 & \textbf{88.18$\:\pm\:$0.4} & 72.29$\:\pm\:$0.2\\
 & \textsc{mpot} & 32.96$\:\pm\:$0.4 & 49.73$\:\pm\:$1.1 & 57.39$\:\pm\:$1.4 & 44.11$\:\pm\:$2.4 & 38.66$\:\pm\:$1.2 & 50.06$\:\pm\:$1.0 & 43.74$\:\pm\:$4.3 & 28.66$\:\pm\:$2.6 & 58.40$\:\pm\:$1.9 & 56.90$\:\pm\:$1.9 & 39.34$\:\pm\:$1.2 & 63.14$\:\pm\:$0.6 & 46.92$\:\pm\:$0.4\\
\midrule\midrule
\multirow{7}{*}{\textsc{1-shot}} & \textsc{s. only} & 43.84$\:\pm\:$1.7 & 66.52$\:\pm\:$3.1 & 77.38$\:\pm\:$0.9 & 50.47$\:\pm\:$2.4 & 53.24$\:\pm\:$2.0 & 61.77$\:\pm\:$1.1 & 56.11$\:\pm\:$1.7 & 37.35$\:\pm\:$1.0 & 71.97$\:\pm\:$1.8 & 68.96$\:\pm\:$0.5 & 46.13$\:\pm\:$2.0 & 73.33$\:\pm\:$2.2 & 58.92$\:\pm\:$0.4\\
 & \textsc{pada} & 52.98$\:\pm\:$0.2 & 63.03$\:\pm\:$1.6 & 78.06$\:\pm\:$2.6 & 51.67$\:\pm\:$5.0 & 56.28$\:\pm\:$0.4 & 64.00$\:\pm\:$1.4 & 58.92$\:\pm\:$3.3 & 43.62$\:\pm\:$1.0 & 74.27$\:\pm\:$4.1 & 68.26$\:\pm\:$3.1 & 54.25$\:\pm\:$1.6 & 78.62$\:\pm\:$0.4 & 62.00$\:\pm\:$0.5\\
 & \textsc{safn} & 31.40$\:\pm\:$3.7 & 49.73$\:\pm\:$4.3 & 62.82$\:\pm\:$2.0 & 48.88$\:\pm\:$2.4 & 45.27$\:\pm\:$0.7 & 57.26$\:\pm\:$2.2 & 42.33$\:\pm\:$1.6 & 29.77$\:\pm\:$2.6 & 63.52$\:\pm\:$3.2 & 56.11$\:\pm\:$3.2 & 37.55$\:\pm\:$0.8 & 67.00$\:\pm\:$1.4 & 49.30$\:\pm\:$0.7\\
 & \textsc{ba3us} & 44.60$\:\pm\:$21.0 & 51.39$\:\pm\:$29.8 & 65.47$\:\pm\:$27.2 & 65.63$\:\pm\:$1.4 & 59.78$\:\pm\:$15.3 & 68.49$\:\pm\:$1.3 & 68.38$\:\pm\:$1.7 & 57.83$\:\pm\:$1.3 & \textbf{82.05$\:\pm\:$1.0} & \textbf{80.78$\:\pm\:$1.1} & \textbf{63.10$\:\pm\:$0.8} & 79.20$\:\pm\:$1.1 & 65.56$\:\pm\:$7.6\\
 & \textsc{ar} & 56.00$\:\pm\:$2.3 & \textbf{78.58$\:\pm\:$1.9} & 82.77$\:\pm\:$2.0 & 68.99$\:\pm\:$0.2 & 68.35$\:\pm\:$1.9 & 77.25$\:\pm\:$1.4 & 69.67$\:\pm\:$1.5 & 51.98$\:\pm\:$1.8 & 78.72$\:\pm\:$1.0 & 76.19$\:\pm\:$0.7 & 55.48$\:\pm\:$2.1 & 82.73$\:\pm\:$1.0 & 70.56$\:\pm\:$0.7\\
 & \textsc{jumbot} & \textbf{61.59$\:\pm\:$1.7} & 76.86$\:\pm\:$3.4 & \textbf{86.45$\:\pm\:$2.1} & \textbf{74.20$\:\pm\:$0.9} & \textbf{73.43$\:\pm\:$3.3} & \textbf{79.85$\:\pm\:$0.3} & \textbf{74.96$\:\pm\:$3.4} & \textbf{62.87$\:\pm\:$0.6} & 81.83$\:\pm\:$0.9 & 78.48$\:\pm\:$2.0 & 61.59$\:\pm\:$2.2 & \textbf{87.34$\:\pm\:$0.2} & \textbf{74.95$\:\pm\:$0.1}\\
 & \textsc{mpot} & 53.97$\:\pm\:$1.3 & 68.78$\:\pm\:$1.7 & 78.04$\:\pm\:$2.1 & 69.24$\:\pm\:$0.4 & 65.88$\:\pm\:$0.5 & 71.42$\:\pm\:$0.7 & 70.31$\:\pm\:$1.0 & 53.03$\:\pm\:$0.7 & 76.88$\:\pm\:$1.3 & 76.52$\:\pm\:$0.4 & 57.39$\:\pm\:$1.7 & 77.95$\:\pm\:$1.4 & 68.28$\:\pm\:$0.2\\
\midrule\midrule
\multirow{7}{*}{\textsc{100-rnd}} & \textsc{s. only} & 43.28$\:\pm\:$1.6 & 68.76$\:\pm\:$1.6 & 77.97$\:\pm\:$1.2 & 53.75$\:\pm\:$1.1 & 55.57$\:\pm\:$2.2 & 63.94$\:\pm\:$0.4 & 58.37$\:\pm\:$0.4 & 39.12$\:\pm\:$0.4 & 75.56$\:\pm\:$1.3 & 69.02$\:\pm\:$0.5 & 43.46$\:\pm\:$0.2 & 75.28$\:\pm\:$2.3 & 60.34$\:\pm\:$0.4\\
 & \textsc{pada} & 50.41$\:\pm\:$0.8 & 67.21$\:\pm\:$1.8 & 79.97$\:\pm\:$1.5 & 56.69$\:\pm\:$1.5 & 53.86$\:\pm\:$1.6 & 63.94$\:\pm\:$1.3 & 60.27$\:\pm\:$2.7 & 40.56$\:\pm\:$1.8 & 78.91$\:\pm\:$1.8 & 72.70$\:\pm\:$1.4 & 53.39$\:\pm\:$2.2 & 80.73$\:\pm\:$0.9 & 63.22$\:\pm\:$0.1\\
 & \textsc{safn} & 47.58$\:\pm\:$0.8 & 67.53$\:\pm\:$0.8 & 77.91$\:\pm\:$0.4 & 56.47$\:\pm\:$1.0 & 58.19$\:\pm\:$0.4 & 65.88$\:\pm\:$0.2 & 59.69$\:\pm\:$0.1 & 43.14$\:\pm\:$1.7 & 75.00$\:\pm\:$0.7 & 69.64$\:\pm\:$1.0 & 50.85$\:\pm\:$0.3 & 76.41$\:\pm\:$0.8 & 62.36$\:\pm\:$0.2\\
 & \textsc{ba3us} & \textbf{62.53$\:\pm\:$2.0} & \textbf{82.09$\:\pm\:$0.8} & \textbf{88.28$\:\pm\:$0.4} & 69.15$\:\pm\:$1.2 & 71.65$\:\pm\:$1.5 & 77.21$\:\pm\:$0.6 & \textbf{75.15$\:\pm\:$1.3} & 58.17$\:\pm\:$1.0 & \textbf{85.92$\:\pm\:$1.3} & 79.86$\:\pm\:$2.1 & \textbf{66.57$\:\pm\:$1.5} & 85.66$\:\pm\:$1.0 & 75.19$\:\pm\:$0.4\\
 & \textsc{ar} & 54.89$\:\pm\:$2.0 & 78.54$\:\pm\:$1.4 & 84.34$\:\pm\:$0.6 & 64.95$\:\pm\:$2.4 & 69.00$\:\pm\:$3.7 & 79.57$\:\pm\:$0.2 & 66.73$\:\pm\:$0.3 & 50.85$\:\pm\:$1.4 & 82.39$\:\pm\:$1.9 & 74.66$\:\pm\:$2.3 & 55.42$\:\pm\:$1.6 & 82.80$\:\pm\:$0.4 & 70.34$\:\pm\:$0.2\\
 & \textsc{jumbot} & 61.07$\:\pm\:$0.9 & 77.87$\:\pm\:$1.4 & 86.01$\:\pm\:$1.3 & \textbf{74.56$\:\pm\:$0.4} & \textbf{76.40$\:\pm\:$1.4} & \textbf{81.54$\:\pm\:$1.7} & 72.60$\:\pm\:$1.2 & \textbf{59.92$\:\pm\:$0.4} & 84.63$\:\pm\:$2.3 & \textbf{81.85$\:\pm\:$1.7} & 64.84$\:\pm\:$1.0 & \textbf{87.64$\:\pm\:$0.7} & \textbf{75.74$\:\pm\:$0.3}\\
 & \textsc{mpot} & 61.59$\:\pm\:$1.2 & 75.56$\:\pm\:$1.7 & 82.59$\:\pm\:$0.6 & 72.48$\:\pm\:$1.0 & 69.77$\:\pm\:$0.9 & 75.41$\:\pm\:$0.5 & 72.64$\:\pm\:$0.9 & 57.67$\:\pm\:$1.6 & 82.02$\:\pm\:$0.6 & 79.80$\:\pm\:$0.5 & 64.64$\:\pm\:$0.1 & 82.60$\:\pm\:$0.5 & 73.06$\:\pm\:$0.3\\
\midrule\midrule
\multirow{7}{*}{\textsc{oracle}} & \textsc{s. only} & 45.43$\:\pm\:$0.9 & 68.91$\:\pm\:$1.4 & 79.53$\:\pm\:$0.3 & 55.59$\:\pm\:$0.7 & 57.42$\:\pm\:$1.2 & 65.23$\:\pm\:$0.8 & 59.32$\:\pm\:$0.7 & 40.80$\:\pm\:$0.9 & 75.80$\:\pm\:$1.2 & 69.88$\:\pm\:$0.9 & 47.20$\:\pm\:$0.9 & 77.31$\:\pm\:$0.1 & 61.87$\:\pm\:$0.3\\
 & \textsc{pada} & 50.53$\:\pm\:$0.7 & 67.45$\:\pm\:$1.6 & 80.14$\:\pm\:$1.4 & 57.30$\:\pm\:$1.9 & 54.47$\:\pm\:$1.7 & 64.55$\:\pm\:$1.1 & 61.07$\:\pm\:$3.0 & 40.94$\:\pm\:$1.6 & 79.55$\:\pm\:$1.4 & 73.09$\:\pm\:$1.5 & 54.63$\:\pm\:$0.9 & 80.93$\:\pm\:$0.6 & 63.72$\:\pm\:$0.3\\
 & \textsc{safn} & 49.57$\:\pm\:$0.3 & 68.55$\:\pm\:$1.0 & 78.26$\:\pm\:$0.2 & 57.91$\:\pm\:$0.3 & 59.29$\:\pm\:$0.5 & 66.81$\:\pm\:$0.5 & 59.87$\:\pm\:$0.7 & 45.29$\:\pm\:$0.7 & 75.98$\:\pm\:$0.6 & 69.08$\:\pm\:$0.6 & 51.68$\:\pm\:$0.8 & 77.29$\:\pm\:$0.5 & 63.30$\:\pm\:$0.2\\
 & \textsc{ba3us} & 63.26$\:\pm\:$1.0 & \textbf{82.75$\:\pm\:$0.9} & \textbf{89.16$\:\pm\:$0.2} & 69.91$\:\pm\:$0.2 & 71.93$\:\pm\:$1.6 & 77.58$\:\pm\:$0.9 & 75.73$\:\pm\:$1.3 & 59.94$\:\pm\:$0.7 & \textbf{86.89$\:\pm\:$0.5} & 80.93$\:\pm\:$0.8 & 66.77$\:\pm\:$1.5 & 86.93$\:\pm\:$0.2 & 75.98$\:\pm\:$0.3\\
 & \textsc{ar} & 57.33$\:\pm\:$1.7 & 79.61$\:\pm\:$1.6 & 86.31$\:\pm\:$0.4 & 69.45$\:\pm\:$0.5 & 71.88$\:\pm\:$0.9 & 79.94$\:\pm\:$0.8 & 70.28$\:\pm\:$1.0 & 53.57$\:\pm\:$0.2 & 83.78$\:\pm\:$1.0 & 77.26$\:\pm\:$0.6 & 59.68$\:\pm\:$1.1 & 83.72$\:\pm\:$0.6 & 72.73$\:\pm\:$0.3\\
 & \textsc{jumbot} & 61.87$\:\pm\:$1.4 & 78.19$\:\pm\:$2.4 & 88.11$\:\pm\:$1.5 & \textbf{77.69$\:\pm\:$0.1} & 76.75$\:\pm\:$0.8 & \textbf{84.15$\:\pm\:$1.3} & \textbf{76.83$\:\pm\:$1.9} & 63.72$\:\pm\:$0.5 & 84.80$\:\pm\:$1.3 & \textbf{81.79$\:\pm\:$0.8} & 64.70$\:\pm\:$1.1 & \textbf{87.17$\:\pm\:$1.7} & 77.15$\:\pm\:$0.4\\
 & \textsc{mpot} & \textbf{64.48$\:\pm\:$1.2} & 80.88$\:\pm\:$3.3 & 86.78$\:\pm\:$0.5 & 76.22$\:\pm\:$0.1 & \textbf{77.95$\:\pm\:$1.3} & 82.59$\:\pm\:$0.7 & 75.18$\:\pm\:$1.3 & \textbf{64.60$\:\pm\:$0.0} & 84.87$\:\pm\:$1.4 & 80.59$\:\pm\:$0.6 & \textbf{67.04$\:\pm\:$0.6} & 86.52$\:\pm\:$1.2 & \textbf{77.31$\:\pm\:$0.5}\\
\end{tabular}}
\caption{Average accuracy of different PDA methods based on different model selection strategies on the 12 tasks of Partial \textsc{office-home}. Average is done over three seeds (2020, 2021, 2022). Best results in \textbf{bold}.}
\label{tab:office_home_results_tasks_with_stds}
\end{table}

%% file: tables/table_visda_comparison_reported.tex
\begin{table}[t!]\centering
\begin{tabular}{c@{\hskip 3pt}|c@{\hskip 3pt}|c@{\hskip 3pt}|c}
\textsc{Algorithm} & SR & RS & Avg\\
\midrule
\textsc{s. only}$^\dagger$ & 45.26 & 64.28 & 54.77\\
\textsc{s. only} (Ours) & 51.86 & 67.11 & 59.48\\
\midrule\midrule
\textsc{pada}$^\dagger$ & 53.53 & 76.50 & 65.02\\
\textsc{pada} (Ours) & 49.34 & 59.81 & 54.57\\
\midrule\midrule
\textsc{safn}$^\dagger$ & 67.65 & - & -\\
\textsc{safn} (Ours) & 56.88 & 68.40 & 62.64\\
\midrule\midrule
\textsc{ba3us}$^\dagger$ & 69.86 & 67.56 & 68.71\\
\textsc{ba3us} (Ours) & 71.77 & 63.56 & 67.67\\
\midrule\midrule
\textsc{ar}$^{\dagger *}$ & 85.30 & 74.82 & 80.06\\
\textsc{ar} (Ours) & 76.33 & 71.36 & 73.85\\
\midrule\midrule
\textsc{jumbot}$^\dagger$ & - & - & -\\
\textsc{jumbot} (Ours) & 90.55 & 77.46 & 84.01\\
\midrule\midrule
\textsc{mpot}$^\dagger$ & - & - & -\\
\textsc{mpot} (Ours) & 87.23 & 86.67 & 86.95\\
\end{tabular}
\caption{Comparison between reported ($\dagger$) accuracies on partial \textsc{visda} from published methods with our implementation using the \textsc{oracle} model selection strategy. * denotes different bottleneck architectures.}
\label{table:visda_comparison_reported}
\end{table}

%% file: tables/table_overral_results_with_50rnd.tex
\begin{table}[t!]\centering
\resizebox{\textwidth}{!}{
\begin{tabular}{c|@{\hskip 3pt}c|@{\hskip 3pt}|c@{\hskip 3pt}|c@{\hskip 3pt}|c@{\hskip 3pt}|c@{\hskip 3pt}|c@{\hskip 3pt}|c@{\hskip 3pt}|c@{\hskip 3pt}|c}
\textsc{Dataset} & \textsc{Method} & \textsc{s-acc} & \textsc{ent} & \textsc{dev} & \textsc{snd} & \textsc{1-shot} & \textsc{50-rnd} & \textsc{100-rnd} & \textsc{oracle}\\
\midrule\midrule
\multirow{7}{*}{\textsc{office-home}} & \textsc{s. only} & 60.38$\pm$0.5& 60.73$\pm$0.2& 60.22$\pm$0.3& 59.55$\pm$0.3& 58.92$\pm$0.4& 60.28$\pm$0.4& 60.34$\pm$0.4& 61.87$\pm$0.3\\
 & \textsc{pada} & 63.08$\pm$0.3& 59.74$\pm$0.5& 52.72$\pm$2.8& 62.36$\pm$0.4& 62.00$\pm$0.5& 63.82$\pm$0.4& 63.22$\pm$0.1& 63.72$\pm$0.3\\
 & \textsc{safn} & 62.09$\pm$0.2& 61.37$\pm$0.3& 62.03$\pm$0.4& 62.59$\pm$0.1& 49.30$\pm$0.7& 62.00$\pm$0.2& 62.36$\pm$0.2& 63.30$\pm$0.2\\
 & \textsc{ba3us} & \textbf{68.32$\pm$1.1}& 73.36$\pm$0.6& 62.25$\pm$7.1& \textbf{75.37$\pm$0.8}& 65.56$\pm$7.6& \textbf{73.22$\pm$0.3}& 75.19$\pm$0.4& 75.98$\pm$0.3\\
 & \textsc{ar} & 65.68$\pm$0.3& 70.58$\pm$0.4& \textbf{64.32$\pm$0.9}& 70.25$\pm$0.2& 70.56$\pm$0.7& 70.26$\pm$0.2& 70.34$\pm$0.2& 72.73$\pm$0.3\\
 & \textsc{jumbot} & 62.89$\pm$0.2& \textbf{74.61$\pm$0.8}& 61.28$\pm$0.1& 72.29$\pm$0.2& \textbf{74.95$\pm$0.1}& 64.95$\pm$0.3& \textbf{75.74$\pm$0.3}& 77.15$\pm$0.4\\
 & \textsc{mpot} & 66.24$\pm$0.1& 64.46$\pm$0.1& 61.37$\pm$0.2& 46.92$\pm$0.4& 68.28$\pm$0.2& 69.90$\pm$0.5& 73.06$\pm$0.3& \textbf{77.31$\pm$0.5}\\
\midrule\midrule
\multirow{7}{*}{\textsc{visda}} & \textsc{s. only} & 55.15$\pm$2.4& 55.24$\pm$3.2& 55.07$\pm$1.2& 55.02$\pm$2.9& 55.72$\pm$2.2& 57.90$\pm$1.1& 58.16$\pm$0.6& 59.48$\pm$0.4\\
 & \textsc{pada} & 47.48$\pm$4.8& 32.32$\pm$4.9& 43.43$\pm$5.3& 56.83$\pm$1.0& 53.15$\pm$2.9& 55.67$\pm$2.5& 54.38$\pm$2.7& 54.57$\pm$2.6\\
 & \textsc{safn} & 58.20$\pm$1.7& 42.83$\pm$6.3& 58.62$\pm$1.3& 44.82$\pm$8.8& 56.89$\pm$2.1& 57.90$\pm$3.3& 59.09$\pm$2.8& 62.64$\pm$1.5\\
 & \textsc{ba3us} & 55.10$\pm$3.7& 65.58$\pm$1.4& 58.40$\pm$1.4& 51.07$\pm$4.3& 64.77$\pm$1.4& 66.66$\pm$2.4& 67.44$\pm$1.2& 67.67$\pm$1.3\\
 & \textsc{ar} & 66.68$\pm$1.0& 64.27$\pm$3.6& \textbf{67.20$\pm$1.5}& 55.69$\pm$0.9& 70.29$\pm$1.7& 71.91$\pm$0.3& 72.60$\pm$0.8& 73.85$\pm$0.9\\
 & \textsc{jumbot} & 60.63$\pm$0.7& 62.42$\pm$2.4& 59.86$\pm$0.6& 77.69$\pm$4.2& \textbf{78.34$\pm$1.9}& 82.85$\pm$2.9& 83.49$\pm$1.9& 84.01$\pm$1.9\\
 & \textsc{mpot} & \textbf{70.02$\pm$2.0}& \textbf{74.64$\pm$4.4}& 61.62$\pm$1.3& \textbf{78.40$\pm$3.9}& 70.96$\pm$3.7& \textbf{86.65$\pm$5.1}& \textbf{86.69$\pm$5.1}& \textbf{86.95$\pm$5.0}\\
\end{tabular}}
\caption{Task accuracy average for the different PDA methods and model selection strategy pairs on Partial Office-Home and Partial VisDA. The average is computed over three difference seeds (2020, 2021, 2022).}
\label{tab:overall_results_50}
\end{table}